%% file: main.tex
\documentclass[10pt,twocolumn,letterpaper]{article}
\usepackage[pagenumbers]{cvpr}
\usepackage[T1]{fontenc}
\usepackage{multirow,colortbl}
\usepackage{cuted}
\usepackage{pifont}
\graphicspath{{figures/}}

\definecolor{cvprblue}{rgb}{0.21,0.49,0.74}
\usepackage[pdfversion=1.7,pagebackref,breaklinks,colorlinks,allcolors=cvprblue]{hyperref}
\hypersetup{pdfauthor={Moxu Duan, Mengfei Gu, Jingwen Fu, Yuwang Wang},pdfsubject={},pdfkeywords={}}
\def\confName{CVPR}
\def\confYear{2027}
\title{Swerve, Don't Wait: In-Flight Action Response for Interactive World Models}
\hypersetup{pdftitle={Swerve, Don't Wait: In-Flight Action Response for Interactive World Models}}
\author{%
  \textbf{Moxu Duan}\textsuperscript{1}\hspace{3.25em}%
  \textbf{Mengfei Gu}\textsuperscript{1}\hspace{3.25em}%
  \textbf{Jingwen Fu}\textsuperscript{2,\textdagger}\hspace{3.25em}%
  \textbf{Yuwang Wang}\rlap{\textsuperscript{3,\textdagger}}\\[8pt]
  \llap{\textsuperscript{1}\,}Beijing Jiaotong University\hspace{1.25em}
  \textsuperscript{2}\,Zhongguancun Academy\hspace{1.25em}
  \textsuperscript{3}\,Tsinghua University\\[2pt]
  {\normalsize\textbf{Project page:} \href{https://paprikkam9.github.io/swerve-project/}{\textup{https://paprikkam9.github.io/swerve-project/}}}%
}

\ifdefined\nolinenumbers\else
  \newenvironment{nolinenumbers}{}{}
\fi

\makeatletter
\patchcmd{\@maketitle}{\vspace*{24pt}}{\vspace*{17pt}}{}{%
  \PackageError{swerve}{Unable to adjust the title-to-author gap}{}}
\patchcmd{\@maketitle}{\vspace*{12pt}}{\vspace*{0pt}}{}{%
  \PackageError{swerve}{Unable to adjust the author-block spacing}{}}
\patchcmd{\@maketitle}{\vskip .5em}{\vskip 1pt}{}{%
  \PackageError{swerve}{Unable to adjust the spacing after the project link}{}}
\makeatother

\begin{document}
\maketitle
\begin{strip}
  \vspace*{-25pt}
  \centering
  \includegraphics[width=\linewidth]{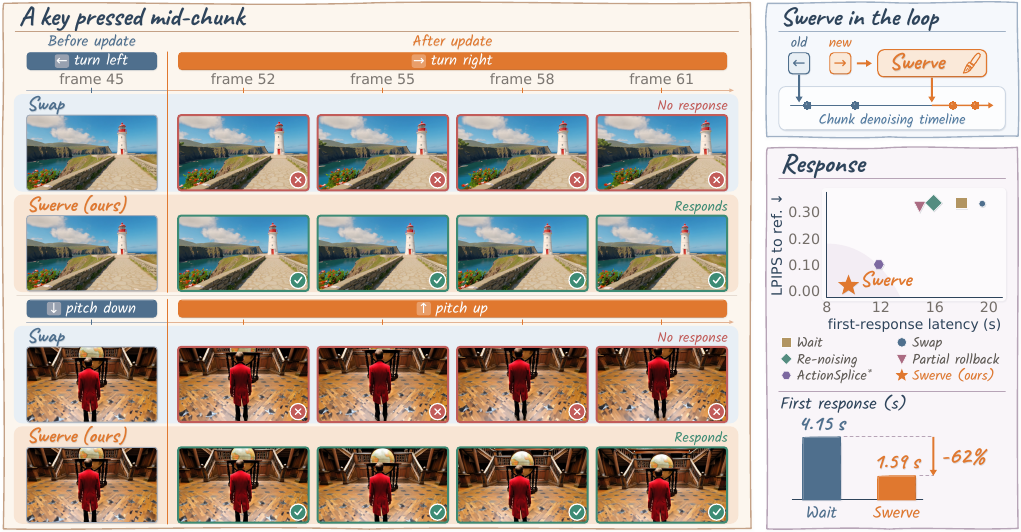}
  \captionsetup{hypcap=false}
  \captionof{figure}{\textbf{Swerve responds to in-flight actions within the current chunk.} \emph{Left.} After a mid-chunk key press, condition swapping (Swap) keeps the old action (\ding{55}) while Swerve follows the new one (\ding{51}). \emph{Top right.} Swerve edits the denoising state without training. \emph{Bottom right.} LPIPS to the full reference vs. first-response latency on HY-WorldPlay, and first-response latency on minWM.}
  \label{fig:teaser}
  \captionsetup{hypcap=true}
\end{strip}
\newskip\FirstPageFootnoteSkip
\FirstPageFootnoteSkip=\skip\footins
\skip\footins=7pt\relax
\enlargethispage{13pt}
\AddToHookNext{shipout/after}{\global\skip\footins=\FirstPageFootnoteSkip}
\begingroup
  \renewcommand{\thefootnote}{\textdagger}
  \footnotetext[1]{Corresponding authors.}
\endgroup

\begin{abstract}
Interactive video world models generate video chunk by chunk, causing a critical latency bottleneck: actions arriving mid-generation (in-flight actions) are delayed until the next chunk. Existing training-free updates fail to effectively alter the current chunk's trajectory, while trained correctors lack generalization across different backbones and novel actions. We uncover denoising inertia as the root cause of this failure: in few-step samplers, the first denoising step largely dictates the action trajectory (persistence), while later steps read little of the action (resistance) and mainly add refinement that is largely shared across actions. Guided by this insight, we propose Swerve, a training-free, plug-and-play denoising-state editor for interactive video world models. Swerve rewrites the first step under the new action, partially retains the reusable refinement from completed steps, and reallocates the remaining steps to preserve the original computational budget. A frame mask further enables precise sub-chunk control, seamlessly handling short key presses and multiple rapid actions within a single chunk. Extensive experiments across diverse mainstream backbones demonstrate its broad applicability: Swerve responds $2.61\times$ and $1.87\times$ earlier than the default waiting strategy on minWM and HY-WorldPlay, respectively, achieving near-perfect fidelity to full regeneration (LPIPS of 0.040 and 0.025) with uncompromised visual quality ($-0.03\%$ VBench score). Furthermore, Swerve is orthogonal to existing training-free accelerators, further boosting HY-WorldPlay's response speedup from $1.87\times$ to $2.70\times$.
\end{abstract}

\section{Introduction}
\label{sec:intro}

Interactive video world models turn a text or image prompt into a virtual world that responds to the actions of users and agents, with applications in game simulation, virtual scene exploration and the training of embodied agents~\cite{diamond,nwm,matrixgame2,yume15,genie3,worldplay,gamengen}. For these models, real-time interaction is the core requirement, and a few hundred milliseconds of input latency already impair controllability~\cite{halbhuber2023latency,liu2022latency,spjut2019latency}. Recent models reach real-time throughput through few-step generation and training-free acceleration~\cite{selfforcing,lightinteraction,causvid}, yet high throughput alone does not make the response prompt~\cite{actionsplice}. These models generate video chunk by chunk, denoising the frames of a chunk together under one action in a few steps, each a single evaluation of the network~\cite{selfforcing,causvid}. Since a user may act at any moment, a new action commonly arrives while a chunk is still being denoised, as an in-flight action~\cite{actionsplice}. This mismatch between the chunk-wise timeline of denoising and the frame-wise timeline of user actions creates a critical latency bottleneck. By default, the model applies the new action only from the next chunk~\cite{yume15,worldplay,minwm}, delaying the response by an entire chunk and rendering sub-chunk interactions, such as short key presses, unrepresentable.

Responding within the current chunk is possible by regenerating it under the new action, which serves as the ideal full reference~\cite{actionsplice}. Regeneration, however, repeats the steps already completed and thus delays the response. Existing in-flight updates therefore aim to respond promptly and faithfully to the new action by reusing the completed steps at no extra sampling cost. Condition swapping simply switches the remaining steps to the new action and expects them to pull the denoising trajectory of the chunk toward it. The frames are ready on time yet still largely follow the old action (\cref{fig:teaser}). Partial rollback and re-noising go further and reopen part of what is already settled, by undoing the last step or by adding fresh noise to the latest prediction before the remaining steps~\cite{sdedit,actionsplice}. Even so, the trajectory remains locked onto the old action. ActionSplice~\cite{actionsplice} instead circumvents the problem by training a corrector network to steer the trajectory toward the full reference. Yet such trained correctors are inherently tied to specific backbones and predefined action transitions, lacking generalization beyond them. The trajectory thus remains uncorrected after switching, stepping back and even training. Is it truly this hard to turn, or is it being turned in the wrong place?

\begin{figure}[t]
  \centering
  \includegraphics[width=\linewidth]{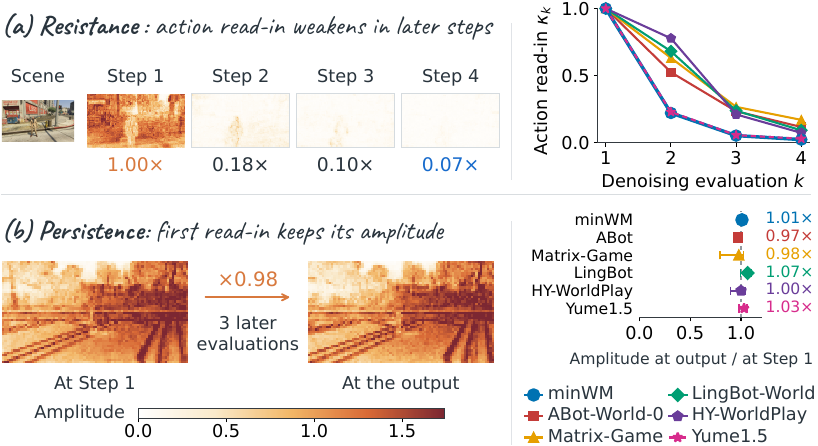}
  \caption{\textbf{Denoising inertia.} The action read-in of a step is the prediction shift from switching the action at that step alone. (a) Resistance. Read-in of each step relative to the first, for one event (left) and six models (right). (b) Persistence. The first read-in at step 1 and at the output (left), and their amplitude ratio (right).}
  \label{fig:denoising-inertia}
\end{figure}

To answer this question, we examine how a few-step sampler reads the action across denoising steps. For each step, its action read-in quantifies the prediction shift induced by switching the action condition at that step alone (\cref{fig:denoising-inertia}). Across six diverse, mainstream interactive world models, the read-in reveals a property that we call \emph{denoising inertia}, with two complementary sides. Specifically, \emph{resistance} means that the first step reads the action in full, while later steps read progressively less of it, dropping to just 2--17\% at the final step. This explains why the remaining steps cannot pull an established trajectory toward a new action. \emph{Persistence} means that what the first step reads is preserved in amplitude and reaches the output at 97--107\%, so the old action largely dictates the trajectory of the chunk. The course of a chunk is thus set by its first step, and existing updates, which all act after this step, work against its inertia. The later steps instead mainly add refinement that is largely shared across actions, and we find that the refinement completed under the old action still points the way the new action would take it.

Guided by this insight, we propose \textbf{Swerve}, a training-free, plug-and-play denoising-state editor for interactive video world models (\cref{fig:overview}). By denoising inertia, a partly denoised chunk follows a course set by its first step under the action, and carries refinement added by later steps that read far less of it. When an in-flight action arrives, Swerve changes the course and keeps the refinement, in three synergistic operations. \textbf{Rewrite} recomputes the first step under the new action to steer the trajectory from its origin. Since the input of this step, the initial noise and the history, does not depend on the action, the rewrite is exact and needs no training. Turning at the origin, however, sets aside the refinement already completed and adds one step. \textbf{Retain} recovers the refinement. It carries part of what the completed steps accumulated under the old action over to the rewritten prediction, allowing the new trajectory to resume near where the interrupted one paused. \textbf{Reallocate} recovers the step. With the refinement back in place, the step that would follow the arrival has less left to add, so it is reassigned to the rewrite, and the chunk completes with its original number of steps. A frame mask further confines the edit to the latent frames from which the action applies, enabling precise sub-chunk control that seamlessly handles short key presses and multiple rapid actions within a single chunk. Swerve changes only the sampling loop and leaves the weights and architecture of the world model untouched, so existing chunk-wise few-step models can adopt it directly.

Since frames can be ready on time without following the new action, we measure the response by the first frame whose estimated camera motion follows it~\cite{vipe}. On minWM and HY-WorldPlay, Swerve responds $2.61\times$ and $1.87\times$ earlier than waiting for the next chunk, achieving near-perfect fidelity to the full reference (LPIPS of 0.040 and 0.025) with uncompromised visual quality ($-0.03\%$ VBench score) across all tested action transitions. Its frame mask further lets the action take effect from a given frame within a chunk, which we demonstrate with short key presses and a second action arriving within the same chunk. Furthermore, Swerve is orthogonal to existing training-free accelerators~\cite{teacache,lightinteraction,sageattention}, which reduce the cost of each chunk while new actions still wait for the next chunk, and combining them with Swerve further boosts the response speedup on HY-WorldPlay from $1.87\times$ to $2.70\times$.

Our contributions are summarized as follows.

\begin{itemize}

  \item We uncover \emph{denoising inertia} in few-step world models as the root cause of in-flight update failures, and show that it also determines which completed refinement remains reusable across action changes.

  \item We propose \textbf{Swerve}, a training-free, plug-and-play denoising-state editor that lets world models respond to in-flight actions within the current chunk and within the original computational budget.

  \item Experiments on minWM and HY-WorldPlay show that Swerve gives the earliest and most faithful response among in-flight updates with uncompromised visual quality, and complements training-free accelerators.

\end{itemize}

\section{Related work}
\label{sec:related}

\noindent\textbf{Interactive video world models.} Interactive video world models simulate an environment as video that unfolds in response to actions, from simulators of specific games~\cite{gamengen,diamond} to general world models~\cite{genie3,matrixgame2,matrixgame3,worldplay,yume15,lingbotworld,minwm,abotworld0,reworld,interactiveworldsurvey}. They take actions as camera poses~\cite{minwm,reworld}, keyboard and mouse signals~\cite{diamond,matrixgame2,matrixgame3,yume15,abotworld0} or both~\cite{worldplay,lingbotworld}, and keep the history in a causal cache~\cite{causvid,selfforcing,matrixgame2,lingbotworld} or in a memory retrieved by camera pose~\cite{worldmem,contextasmemory,worldplay,reworld}. HY-WorldPlay~\cite{worldplay}, for example, takes discrete keys together with camera poses and rebuilds the context of each chunk from frames selected by camera geometry. Real-time systems among them generate chunk by chunk with few-step causal generators~\cite{causvid,selfforcing,causalforcing}. Swerve edits only the denoising state of a chunk and thus applies across these designs.

\noindent\textbf{In-flight action updates.} Prompt switching and event control let the condition change during generation, by rebuilding the cache, reorganizing the memory or training for smooth transitions~\cite{longlive,infinityrope,echoforcing,anchorforcing,deltaforcing,viskoorbis}. LongLive~\cite{longlive}, for instance, refreshes its cached states when the prompt switches. These methods change only the chunks after a switch, so the response still waits for the next chunk. ActionSplice~\cite{actionsplice} first studies how to update the current chunk, taking the chunk regenerated under the new action as the full reference. Its training-free baselines transfer techniques from diffusion editing, continuing sampling under the new condition~\cite{promptmixing} or restarting the remaining steps from an earlier state or from a noisier one obtained by re-noising~\cite{sdedit,repaint,diffusionreroll}. ActionSplice itself trains a corrector for each backbone and type of edit. Swerve responds earlier and more faithfully than these updates without training, down to individual latent frames.

\noindent\textbf{Training-free generation acceleration.} Training-free acceleration of video generation, including the streaming generation of interactive world models, computes attention efficiently~\cite{sageattention,spargeattn,sparsevideogen}, manages the history that each chunk attends to~\cite{lightinteraction,worldkv}, or reuses outputs across evaluations~\cite{teacache,pab,fastercache,magcache,sencache,arcache,surpriseforcing}. Among them, SageAttention~\cite{sageattention} computes attention in low precision, Light Interaction~\cite{lightinteraction} prunes the history by camera-pose similarity and scene dynamics, and TeaCache~\cite{teacache} reuses the output of an evaluation when its timestep-modulated input changes little. These methods reduce the cost of each chunk, yet an action that arrives during a chunk still waits for the next one. Swerve is orthogonal to them, as it decides which evaluations follow an arrival and leaves how each is computed to the accelerator.

\section{Method}
\label{sec:method}

\subsection{Formulation}
\label{sec:method-formulation}

\noindent\textbf{Chunk-wise few-step sampling.} An interactive world model generates video chunk by chunk~\cite{causvid,selfforcing}. Each chunk of $N$ latent frames is conditioned on the history $\mathcal H$, the previously generated frames that the model attends to through a KV cache or a retrieved memory, and on an action sequence $A=(a_1,\dots,a_N)$ with one action per latent frame. A latent frame is thus the finest unit at which the model takes actions, and frames below refer to latent frames. A few-step sampler~\cite{consistencymodels,causvid,selfforcing} denoises a chunk in $K$ steps. Starting from the initial noise $x_{t_1}\sim\mathcal N(0,I)$, it evaluates the network once per step, at the decreasing noise levels $t_1>\dots>t_K$. With the network written as a denoiser $D_\theta$ that returns a data prediction~\cite{edm,dpmsolverpp}, the $k$-th evaluation and the move to the next noise level are
\begin{equation}
  \hat x_k=D_\theta(x_{t_k},t_k;\mathcal H,A),\; x_{t_{k+1}}=\alpha_{t_{k+1}}\hat x_k+\sigma_{t_{k+1}}\hat\epsilon_k,
  \label{eq:method-sampling}
\end{equation}
where $(\alpha_t,\sigma_t)$ is the noise schedule~\cite{vdm} and $\hat\epsilon_k$ is drawn afresh in consistency sampling~\cite{consistencymodels} or implied by $x_{t_k}$ and $\hat x_k$ in DDIM and rectified-flow sampling~\cite{ddim,rectifiedflow}. The last prediction $\hat x_K$ is the generated chunk, and the noisy latents and predictions of \cref{eq:method-sampling} form its denoising trajectory.

\noindent\textbf{In-flight actions.} A user may issue a new action while the current chunk is being sampled, so that its action sequence changes from $A^-$ to $A^+$ after $r$ of the $K$ evaluations, with $1\le r\le K-1$. The new action is then an in-flight action, and $r$ is its receipt step~\cite{actionsplice}. The number $m$ of leading frames on which $A^-$ and $A^+$ agree locates the change on the video timeline (\cref{fig:overview}). In whole-chunk response, $m=0$. In frame-accurate response, $m>0$, and $A^+$ takes effect from frame $m+1$ on, where the frame mask $M$ is one, while $M$ is zero on the first $m$ frames.

\noindent\textbf{Problem.} Let $\hat x_k^-$ and $\hat x_k^+$ be the $k$-th predictions when the chunk is sampled entirely under $A^-$ and $A^+$, and $\Delta_k=\hat x_k^+-\hat x_k^-$ their action difference. The full reference $x^\star$ is the chunk sampled under $A^+$ from the start with the same initial noise, history and noise draws, with the first $m$ frames reset to their predictions under $A^-$ after every evaluation, so that $x^\star=\hat x_K^+$ in whole-chunk response. When the action arrives, $\hat x_1^-,\dots,\hat x_r^-$, the initial noise, the history and the noise of the sampler form the denoising state $\mathcal S_r$. An update rule $\mathcal U$ maps $\mathcal S_r$ and $A^+$ to a chunk, and the in-flight action update problem is
\begin{equation}
  \min_{\mathcal U}\ \big\|\mathcal U(\mathcal S_r,A^+)-x^\star\big\|,
  \label{eq:method-problem}
\end{equation}
solved with the world model frozen, its noise levels unchanged and, as far as possible, no more evaluations than ordinary sampling. Sampling again under $A^+$ attains $x^\star$ but repeats the $r$ completed evaluations, while the evaluations that remain can turn the trajectory toward $A^+$ only as far as they read the new action.

\subsection{Analysis}
\label{sec:method-analysis}
\label{sec:method-inertia}

\noindent\textbf{Action read-in.} To locate where a few-step sampler reads the action, we sample a chunk under $A^+$ and let only its $k$-th evaluation receive $A^-$. The prediction of that evaluation changes by
\begin{equation}
  \kappa_k=D_\theta(x_{t_k}^+,t_k;\mathcal H,A^-)-D_\theta(x_{t_k}^+,t_k;\mathcal H,A^+),
  \label{eq:method-read-in}
\end{equation}
the action read-in of evaluation $k$, and continuing the sampler under $A^+$ shows how much of it reaches the output. Across six few-step world models that differ in action conditioning, memory and sampler, the read-in shows two properties, which together form denoising inertia (\cref{fig:denoising-inertia}). Under \textbf{resistance}, the read-in weakens in later evaluations, and that of the last one falls to 2--17\% of the first in median. Under \textbf{persistence}, the first read-in, which equals $\Delta_1$ up to sign, reaches the output at 97--107\% of its amplitude. Prior studies find that layout and motion form in the early steps of many-step samplers under a condition fixed before sampling~\cite{ediffi,prompt2prompt,ac3d,baherwanimotion}. Early forming, however, does not tell whether later evaluations can still redirect a chunk once its action changes, which is the question that in-flight actions raise. Denoising inertia answers it by intervention and separates two causes that an aggregate sensitivity conflates, namely that later evaluations read little of the action and that what the first evaluation read is not washed out afterwards. The course of a chunk is thus set by its first evaluation alone, and in a few-step sampler this is a single evaluation that can be recomputed.

\noindent\textbf{Reusable refinement.} Inertia also bears on what the later evaluations contribute. Let $u^-=\hat x_r^--\hat x_1^-$ and $u^+=\hat x_r^+-\hat x_1^+$ be the refinement that the evaluations up to step $r$ add after the first one under $A^-$ and $A^+$. The two refinements differ only by
$u^+-u^-=\Delta_r-\Delta_1,$
the action difference added after the first evaluation. Resistance suggests that the two refinements largely agree, which we verify directly since they arise on different trajectories. At $r=2$ on minWM and HY-WorldPlay, the median cosine similarity $c$ between $u^-$ and $u^+$ is 0.90 and 0.97, while the old refinement is larger, with a median ratio $s=\|u^-\|_2/\|u^+\|_2$ of 1.60 and 1.80. The old refinement thus points the way the new one would but exceeds it in amplitude, so it remains useful under the new action in part. Supplementary material~\ref{sec:supp-analysis} extends both analyses to more models, actions and receipt steps.

\subsection{Swerve}
\label{sec:method-swerve}

\begin{figure*}[t]
  \centering
  \includegraphics[width=\textwidth]{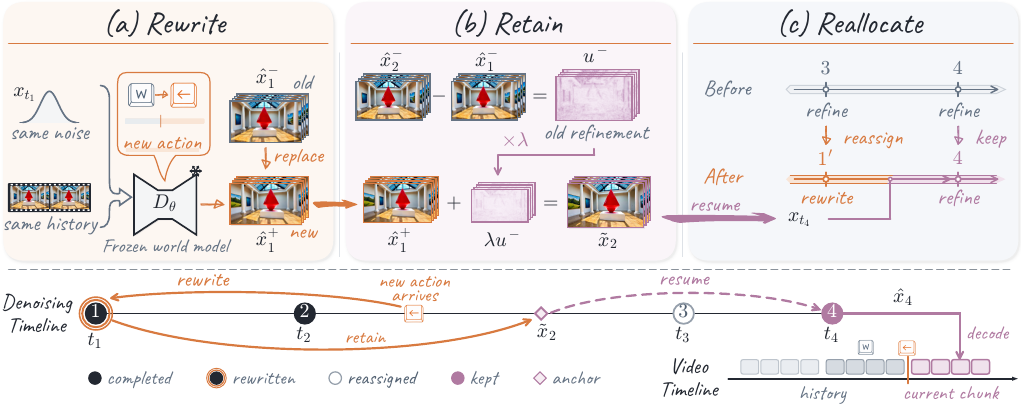}
  \caption{\textbf{Swerve overview.} (a) Rewrite recomputes the first evaluation under the new action from the same initial noise and history. (b) Retain adds a fraction $\lambda$ of the old refinement to the rewritten prediction to form the anchor. (c) Reallocate reassigns the evaluation after the arrival to the rewrite and resumes sampling from the anchor. In the example, the action arrives after the second of $K=4$ evaluations.}
  \label{fig:overview}
\end{figure*}

By \cref{sec:method-analysis}, a partly denoised chunk follows a course set by its first evaluation under $A^-$, and carries refinement added by the later evaluations and largely shared by both actions. Swerve edits the denoising state $\mathcal S_r$ to change the course and keep the refinement, in three synergistic operations, Rewrite, Retain and Reallocate (\cref{fig:overview}).

\noindent\textbf{Rewrite.} Since the course is set by the first evaluation, Swerve turns the trajectory at its start, recomputing the first evaluation under $A^+$ and replacing $\hat x_1^-$ with the result (\cref{fig:overview}a). The input of this evaluation, the initial noise $x_{t_1}$ and the history $\mathcal H$, depends neither on the action nor on the receipt step, whereas the inputs of later evaluations carry the predictions made under $A^-$. The rewrite therefore yields exactly $\hat x_1^+$, the first prediction of the full reference. It requires no training, only the initial noise kept from the start of sampling, and costs one evaluation.

\noindent\textbf{Retain.} The rewritten prediction lacks the refinement that the interrupted trajectory completed after its first evaluation. Since the old refinement $u^-$ points the way the new one would (\cref{sec:method-analysis}), Swerve adds a fraction $\lambda\in[0,1]$ of it to the rewrite and forms the anchor
\begin{equation}
  \tilde x_r=\hat x_1^++\lambda u^-
  \label{eq:method-anchor}
\end{equation}
(\cref{fig:overview}b). The anchor deviates from the prediction of the full reference at the same step by $\lambda u^--u^+$, whose amplitude relative to the refinement $u^+$ that it should recover is
\begin{equation}
  \frac{\|\tilde x_r-\hat x_r^+\|}{\|u^+\|}=\sqrt{1-2\lambda sc+\lambda^2s^2}.
  \label{eq:method-anchor-deviation}
\end{equation}
The rewrite alone, with $\lambda=0$, leaves all of $u^+$ out, while full retention, with $\lambda=1$, also carries the excess amplitude of $u^-$, and \cref{eq:method-anchor-deviation} is smallest in between, at $\lambda=c/s$. With the median $c$ and $s$, this minimum lies at 0.56 and 0.54 on minWM and HY-WorldPlay, and $\lambda=\tfrac12$ leaves 0.45 and 0.25 of $u^+$, close to this minimum, against 0.82 and 0.87 under full retention. Partial retention thus recovers the useful refinement without its excess. Unlike the interrupted prediction, whose deviation $-\Delta_r$ carries the whole difference $\Delta_1$, the anchor carries $\Delta_1$ only through the change $\Delta_r-\Delta_1$.

\noindent\textbf{Reallocate.} The rewrite adds one evaluation to the chunk, which would delay its delivery. By resistance, the evaluation that would follow the arrival reads less of the action than the first one, and the anchor already carries the new action together with part of the refinement that this evaluation would add. Swerve therefore reassigns this evaluation, at $t_{r+1}$, to the rewrite (\cref{fig:overview}c) and moves the anchor to $t_{r+2}$ by \cref{eq:method-sampling}, with $\tilde x_r$ in place of the prediction and the noise of the interrupted sampler. When the noise is drawn afresh, this is the noise drawn for the transition into $t_{r+2}$, and when the sampler is deterministic, it is the noise prediction $\hat\epsilon_r$ of the last completed evaluation. Sampling then resumes under $A^+$ and continues to the final evaluation, so the chunk ends at its native last noise level with as many evaluations as ordinary sampling. When the action arrives just before the last evaluation, no evaluation remains to be reassigned, and the rewrite is followed by that evaluation, one more than ordinary sampling.

\subsection{Frame-accurate response}
\label{sec:method-frame-accurate}

Frame-accurate response requires the first $m$ frames to keep their content while the new actions apply to the frames after them. Since the rewrite reads the whole per-frame sequence $A^+$ in one evaluation, wherever the change lies, Swerve meets this requirement by applying the frame mask $M$ to the anchor alone,
\begin{equation}
  \tilde x_r=M\odot\big(\hat x_1^++\lambda u^-\big)+(1-M)\odot\hat x_r^-,
  \label{eq:method-frame-anchor}
\end{equation}
which reduces to \cref{eq:method-anchor} when $M=\mathbf 1$. The first $m$ frames of the anchor thus keep the prediction of the interrupted trajectory, and reallocation and the remaining evaluations proceed unchanged. The sequence $A^+$ may also differ from $A^-$ on only a few frames, as in a short key press after which the old action resumes. The rewrite then reads all of these changes in its single evaluation, and $M$ covers every frame after the first change.

A user may also press keys in quick succession, so a second action can arrive while the edited chunk is still being sampled. Swerve then treats the edited trajectory as the interrupted one. When a sequence $A^{++}$ replaces $A^+$ at receipt step $r'$, Swerve rewrites the first evaluation under $A^{++}$ and forms the new anchor by \cref{eq:method-frame-anchor}, with the rewritten prediction $\hat x_1^{++}$, the latest prediction $\hat x_{r'}$ of the edited trajectory and its anchor $\tilde x_r$ in place of $\hat x_1^+$, $\hat x_r^-$ and $\hat x_1^-$, and with the mask $M'$ of the frames after those on which $A^+$ and $A^{++}$ agree. The full reference then uses $A^{++}$ in place of $A^+$.

\section{Experiments}
\label{sec:experiments}

This section examines whether Swerve helps interactive world models respond to in-flight actions promptly and faithfully, from whole chunks down to latent frames, and where its gain comes from. Specifically, the evaluation is designed to answer the following research questions.

\begin{enumerate}

  \item \textbf{Q1 (Responsiveness):} Does Swerve make an in-flight action take effect sooner and closer to the full reference than existing updates, without lowering visual quality and also for unseen actions?

  \item \textbf{Q2 (Frame accuracy):} Can Swerve make an action take effect from a given frame and carry several changes within one chunk?

  \item \textbf{Q3 (Ablation and complementarity):} How much does each operation of Swerve contribute, and does its gain add to that of accelerating each evaluation?

\end{enumerate}

\begingroup
\captionsetup{hypcap=false}
\begin{table*}[t]
\captionof{table}{\textbf{Whole-chunk response on minWM and HY-WorldPlay.} Swerve responds earliest and most faithfully at the visual quality of the full reference. Pixel, First: end-to-end pixel-ready, first-response latency. RT: real-time latency, \cref{eq:rt-latency}. Next: LPIPS of the next chunk. Unseen actions: outside the training of ActionSplice. $^\dagger$: full reference (gray). \textsuperscript{*}: trained locally with the official procedure.}
  \label{tab:main-results}
  \centering
  \begingroup
  \small
  \renewcommand{\arraystretch}{0.95}
  \setlength{\aboverulesep}{0.25ex}
  \setlength{\belowrulesep}{0.35ex}
  \setlength{\tabcolsep}{2.1pt}
  \begin{tabular}{@{}>{\columncolor{white}[0pt][\tabcolsep]}llccccccccc>{\columncolor{white}[\tabcolsep][0pt]}c@{}}
    \toprule
    \multirow{2}{*}{\textbf{Model}} & \multirow{2}{*}{\textbf{Method}}
    & \multicolumn{4}{c}{\textbf{Latency}}
    & \multicolumn{3}{c}{\textbf{Response quality}}
    & \multicolumn{1}{c}{\textbf{Visual quality}}
    & \multicolumn{2}{c}{\textbf{Unseen actions}} \\
    \cmidrule(lr){3-6}\cmidrule(lr){7-9}\cmidrule(lr){10-10}\cmidrule(l){11-12}
    & & Pixel (ms)$\downarrow$ & First (ms)$\downarrow$ & RT (ms)$\downarrow$ & Speedup$\uparrow$
    & LPIPS$\downarrow$ & PSNR$\uparrow$ & Next$\downarrow$
    & VBench (\%)$\uparrow$ & LPIPS$\downarrow$ & First (ms)$\downarrow$ \\
    \midrule
    \rowcolor{black!10}
     & \textbf{Full$^\dagger$} & 2103.0 & 3291.8 & 1250.0 & 1.26$\times$ & \textemdash & \textemdash & \textemdash & 86.02 & \textemdash & 2635.4 \\ %
     & Wait & 3864.0 & 4150.8 & 1812.5 & 1.00$\times$ & 0.1292 & 17.48 & 0.3396 & \textbf{86.11} & 0.1968 & 3140.5 \\ %
     & Swap & \underline{1366.3} & 4135.8 & 1812.5 & 1.01$\times$ & 0.1290 & 17.48 & 0.3275 & 86.08 & 0.1966 & \underline{3139.9} \\ %
     & Partial rollback & 1734.9 & 4490.2 & 2062.5 & 0.93$\times$ & 0.1271 & 17.71 & 0.3092 & 86.07 & \underline{0.1940} & 3355.6 \\ %
     & Re-noising & 1367.0 & 4117.0 & 1812.5 & 1.01$\times$ & 0.1332 & 17.49 & 0.3306 & \underline{86.10} & 0.1960 & 3140.1 \\ %
     & ActionSplice\textsuperscript{*} & 1370.8 & \underline{2309.1} & \underline{1062.5} & \underline{1.81}$\times$ & \underline{0.0946} & \underline{18.87} & \underline{0.2556} & 85.79 & 0.1969 & 3144.5 \\ %
    \multirow{-7}{*}{\textbf{minWM}} & \textbf{Swerve (ours)} & \textbf{1365.4} & \textbf{1589.9} & \textbf{750.0} & \textbf{2.61}$\times$ & \textbf{0.0404} & \textbf{22.97} & \textbf{0.0938} & 85.99 & \textbf{0.0568} & \textbf{1206.1} \\ %
    \midrule
    \rowcolor{black!10}
     & \textbf{Full$^\dagger$} & 13696.8 & 13819.6 & 750.0 & 1.30$\times$ & \textemdash & \textemdash & \textemdash & 86.31 & \textemdash & 9203.1 \\ %
     & Wait & 17805.2 & 17951.1 & 1125.0 & 1.00$\times$ & 0.3395 & 13.56 & 0.5184 & \underline{86.17} & 0.3605 & 11898.3 \\ %
     & Swap & \textbf{9466.1} & 19512.1 & 1125.0 & 0.92$\times$ & 0.3367 & 13.61 & 0.4921 & 86.04 & 0.3586 & 12041.1 \\ %
     & Partial rollback & 10605.8 & 14873.1 & 1020.8 & 1.21$\times$ & 0.3277 & 14.55 & 0.4236 & 85.96 & \underline{0.3472} & \underline{10747.4} \\ %
     & Re-noising & 9483.6 & 15911.4 & 958.3 & 1.13$\times$ & 0.3400 & 13.66 & 0.4891 & 85.91 & 0.3593 & 12046.9 \\ %
     & ActionSplice\textsuperscript{*} & 9488.8 & \underline{11849.5} & \underline{604.2} & \underline{1.51}$\times$ & \underline{0.1053} & \underline{20.48} & \underline{0.3947} & 84.85 & 0.3610 & 11639.0 \\ %
    \multirow{-7}{*}{\shortstack[l]{\textbf{HY-}\\\textbf{WorldPlay}}} & \textbf{Swerve (ours)} & \underline{9472.3} & \textbf{9605.8} & \textbf{416.7} & \textbf{1.87}$\times$ & \textbf{0.0248} & \textbf{27.11} & \textbf{0.0576} & \textbf{86.21} & \textbf{0.0319} & \textbf{4926.3} \\ %
    \bottomrule
  \end{tabular}
  \endgroup
\par
\captionof{table}{\textbf{Frame-accurate response on HY-WorldPlay.} Swerve leads on the suffix in every setting. Single boundary: the old action is kept on the first $m\in\{1,2,3\}$ latent frames. Short press: the new action applies to latent frames 2 and 3 only. Second update: another action arrives after three evaluations of a chunk first updated after one. Detected events: number of events with a response.}
  \label{tab:frame-response}
  \centering
  \begingroup
  \small
  \renewcommand{\arraystretch}{0.95}
  \setlength{\aboverulesep}{0.25ex}
  \setlength{\belowrulesep}{0.35ex}
  \setlength{\tabcolsep}{2.4pt}
  \begin{tabular}{@{}>{\columncolor{white}[0pt][\tabcolsep]}lcccccccccc>{\columncolor{white}[\tabcolsep][0pt]}c@{}}
    \toprule
    \multicolumn{1}{c}{\multirow{2}{*}[-4pt]{\textbf{Method}}}
    & \multicolumn{5}{c}{\textbf{Single boundary}}
    & \multicolumn{3}{c}{\textbf{Short press}}
    & \multicolumn{3}{c}{\textbf{Second update}} \\
    \cmidrule(lr){2-6}\cmidrule(lr){7-9}\cmidrule(l){10-12}
    \multicolumn{1}{c}{} & \shortstack{Prefix\\LPIPS$\downarrow$}
    & \shortstack{Suffix\\LPIPS$\downarrow$}
    & \shortstack{Suffix\\PSNR$\uparrow$}
    & \shortstack{Boundary\\error$\downarrow$}
    & \shortstack{First\\Response (ms)$\downarrow$}
    & \shortstack{Suffix\\LPIPS$\downarrow$}
    & \shortstack{Suffix\\PSNR$\uparrow$}
    & \shortstack{Detected\\events$\uparrow$}
    & \shortstack{Suffix\\LPIPS$\downarrow$}
    & \shortstack{Suffix\\PSNR$\uparrow$}
    & \shortstack{First\\Response (ms)$\downarrow$} \\
    \midrule
    \rowcolor{black!10}
    \textbf{Full$^\dagger$} & \textemdash & \textemdash & \textemdash & \textemdash & 11372.3 & \textemdash & \textemdash & 54/54 & \textemdash & \textemdash & 14559.5 \\
    Swap & \textbf{0.0055} & 0.2751 & 15.46 & 1.5921 & 17871.8 & 0.3434 & 14.13 & 2/54 & 0.1303 & 17.24 & 17109.3 \\
    Partial rollback & 0.0117 & 0.2065 & 16.08 & 0.5130 & 12537.9 & \underline{0.2955} & \underline{16.12} & \underline{13/54} & \underline{0.0965} & \underline{21.20} & \underline{14989.6} \\
    Re-noising & 0.0198 & 0.2822 & 15.39 & 1.7133 & 17872.1 & 0.3391 & 14.15 & 1/54 & 0.1410 & 17.12 & 17111.3 \\
    ActionSplice\textsuperscript{*} & \underline{0.0098} & \underline{0.0706} & \underline{24.65} & \underline{0.0920} & \underline{7900.9} & 0.3043 & 15.04 & 9/54 & 0.1065 & 18.67 & 16984.8 \\
    \textbf{Swerve (ours)} & 0.0099 & \textbf{0.0227} & \textbf{28.75} & \textbf{0.0769} & \textbf{7546.3} & \textbf{0.0199} & \textbf{28.67} & \textbf{54/54} & \textbf{0.0219} & \textbf{28.46} & \textbf{7588.7} \\
    \bottomrule
  \end{tabular}
  \endgroup
\end{table*}
\endgroup

\subsection{Experimental setup}
\label{sec:exp-setup}

\noindent\textbf{Models.} We evaluate Swerve on two state-of-the-art open-source world models, minWM-Wan-Action2V-1.3B (minWM)~\cite{minwm} and HY-World1.5-Autoregressive-480P-I2V-distill-8B (HY-WorldPlay)~\cite{worldplay}. Both generate $832\times480$ video in chunks of four latent frames.

\noindent\textbf{Scenes and events.} Each backbone is evaluated on 54 scenes, 30 from the test scenes of ActionSplice~\cite{actionsplice} and 24 from the image-text pairs that Light Interaction builds from VBench~\cite{lightinteraction,vbench}, added to enlarge the evaluation. Each scene has three events, in which an in-flight action switches among forward, backward and yaw left~\cite{actionsplice}. One further event per scene switches to an unseen action, one that ActionSplice is not trained on, namely yaw right, lateral translation or pitch. All methods share the prompt, history, initial noise and action sequence of an event, and all experiments below use these 54 scenes with the events above.

\noindent\textbf{Baselines.} We compare Swerve with Wait, the default that applies the new action from the next chunk, three training-free updates, Swap, partial rollback and re-noising~\cite{sdedit,diffusionreroll,actionsplice}, and the trained corrector of ActionSplice~\cite{actionsplice}. As its weights are not public, we train the correctors locally, strictly following the official training procedure and settings, and denote the result by ActionSplice\textsuperscript{*}, whose results are consistent with those reported in its paper (supplementary material~\ref{sec:supp-setup}). The full reference serves as the oracle.

\noindent\textbf{Evaluation metrics.}

\begin{itemize}

  \item \textbf{Response latency.} Latency is measured in two ways. End to end on our GPUs, the pixel-ready latency runs from the key press until the frames generated under the new action are decoded, and the first-response latency until the first frame that follows the new action is played. The real-time latency models interaction where generation keeps pace with playback~\cite{minwm,worldplay}, denoising each chunk while the preceding one plays. A key press at receipt step $r$ then leaves $(1-r/K)L$ of the preceding chunk to play, $L$ being the chunk playback time, and each of the $n$ evaluations a method adds stalls playback by $L/K$. The latency from the key press to the first displayed frame that follows the new action is
  \begin{equation}
    T_{\mathrm{RT}}=\Big(1-\frac{r}{K}\Big)L+\frac{n}{K}L+\tau,
    \label{eq:rt-latency}
  \end{equation}
  where $\tau$ is the video time from the start of the current chunk to that frame. Following the response detection of WorldMark~\cite{worldmark}, whether a frame follows the action is judged from its camera motion estimated with ViPE~\cite{vipe}, and a manual check of 100 sampled events confirms all of these judgments. Speedup is the ratio of the end-to-end first-response latency of Wait, the default of current interactive world models, to that of a method.

  \item \textbf{Response quality.} LPIPS~\cite{lpips} and PSNR are computed against the full reference. For a change within a chunk, the frames before and after it are scored separately as prefix and suffix. The boundary error is the normalized squared error between the change across the boundary, from one latent frame to the next, and the same change in the full reference~\cite{actionsplice}.

  \item \textbf{Visual quality.} We report the mean score of VBench~\cite{vbench} as the final result.

\end{itemize}

\noindent\textbf{Implementation details.} Both backbones use $K=4$ evaluations per chunk, and the new action arrives after $r=2$ of them in the main experiments. Swerve uses $\lambda=\tfrac12$ on both backbones without tuning. All latencies include decoding and are measured on NVIDIA RTX PRO 6000 Blackwell GPUs. Supplementary material~\ref{sec:supp-setup} details the setup, \ref{sec:supp-receipt-step} sweeps the receipt step $r$, \ref{sec:supp-backbones} reports further backbones, and \ref{sec:supp-realtime} details the evaluation of the real-time latency in \cref{eq:rt-latency}.

\subsection{Whole-chunk response}
\label{sec:exp-whole-chunk}

\noindent\textbf{Main results.} As shown in \cref{tab:main-results}, Swerve responds earliest and most faithfully on both backbones. It responds $2.61\times$ and $1.87\times$ earlier than Wait, with a real-time latency of 750.0 and 416.7 ms against 1812.5 and 1125.0 ms for Wait (\cref{fig:camera-response}), and its LPIPS of 0.040 and 0.025 is 57.3\% and 76.4\% below that of ActionSplice\textsuperscript{*}, the strongest baseline. The training-free baselines respond at most $1.21\times$ earlier than Wait, although Swap and re-noising have frames ready as early as Swerve, and their LPIPS stays at the level of Wait, so their chunks still follow the old action.

In the next chunk, the LPIPS of Swerve stays at 0.094 and 0.058, while that of ActionSplice\textsuperscript{*} rises to 0.256 and 0.395 as its error enters the history, and supplementary material~\ref{sec:supp-consecutive-updates} tracks this error over five consecutive updates. The VBench scores of Swerve, 85.99 and 86.21, stay close to the 86.02 and 86.31 of the full reference.

\begin{nolinenumbers}
\noindent
\begin{minipage}[b]{\columnwidth}
  \centering
  \includegraphics[width=\linewidth]{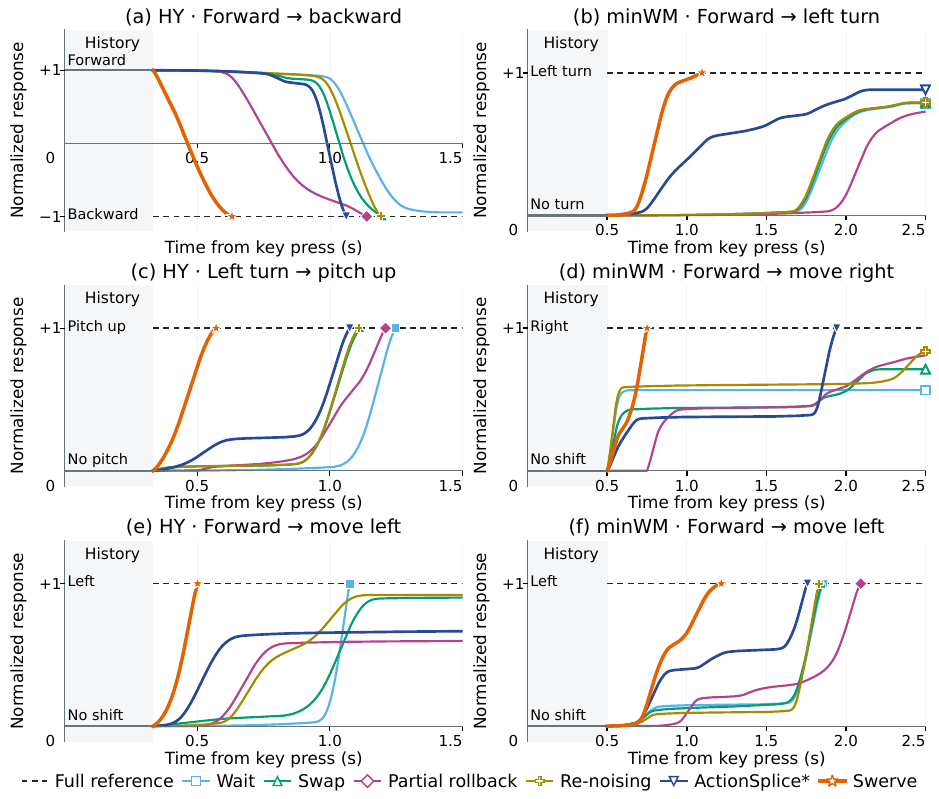}
  \captionsetup{hypcap=false}
  \captionof{figure}{\textbf{Response in real-time interaction.} Camera motion of each update after the key press, timed as in \cref{eq:rt-latency} and normalized to the full reference (dashed). (c)--(f): unseen actions.}
  \label{fig:camera-response}
\end{minipage}
\par
\end{nolinenumbers}

\noindent\textbf{Unseen actions.} Swerve needs no training and thus applies to any action that the backbone accepts. On unseen actions, ActionSplice\textsuperscript{*} falls back to the LPIPS of Swap and the first response of Wait, whereas Swerve keeps an LPIPS of 0.057 and 0.032 and responds in less than half the latency of Wait.

\subsection{Frame-accurate response}
\label{sec:exp-frame-accurate}

\cref{tab:frame-response} evaluates action changes inside a chunk on HY-WorldPlay at the granularity of latent frames, in three settings, a single boundary, a short press and a second update. With the frame mask applied to every update, the prefix LPIPS stays below 0.02, so the updates differ only after the change. At a single boundary, Swerve attains the lowest suffix LPIPS and boundary error, 0.023 and 0.077 against 0.071 and 0.092 for ActionSplice\textsuperscript{*}, which is trained for this case, and responds first. As its rewrite reads the whole per-frame action sequence, Swerve also responds to a short press of two latent frames in all 54 events, against at most 13 for the baselines, and to a second action in the same chunk about twice as early as the best baseline (\cref{fig:response-frames}).

\begingroup
\captionsetup{hypcap=false}
\begin{figure*}[t]
\centering
  \includegraphics[width=\textwidth]{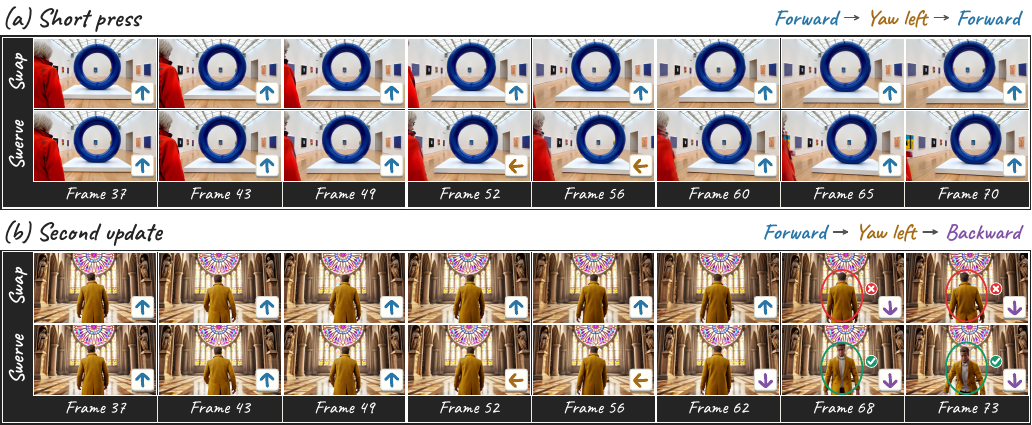}
  \caption{\textbf{Qualitative comparison on HY-WorldPlay.} Swap and Swerve under (a) a short yaw-left press during forward motion and (b) a second update from yaw left to backward. Arrows give the action of each frame, and marks show whether the frame follows it.}
  \label{fig:response-frames}
\par
\centering
  \begin{minipage}[t]{\columnwidth}
\captionof{table}{\textbf{Ablation of the three operations of Swerve on HY-WorldPlay.} The operations are added to Swap one at a time. Rw, Rt, Ra: Rewrite, Retain, Reallocate. Evals: evaluations after the arrival. Pixel, First: pixel-ready, first-response latency.}
  \label{tab:ablation}
\centering
  \begingroup
  \small
  \renewcommand{\arraystretch}{0.95}
  \setlength{\aboverulesep}{0.25ex}
  \setlength{\belowrulesep}{0.35ex}
  \setlength{\tabcolsep}{1.5pt}
  \begin{tabular*}{\columnwidth}{@{\extracolsep{\fill}}lccccccc@{}}
    \toprule
    \multirow{2}{*}{Variant} & \multicolumn{3}{c}{Components} & \multirow{2}{*}{Evals} & \multirow{2}{*}{LPIPS$\downarrow$} & \multicolumn{2}{c}{Latency (ms)$\downarrow$} \\
    \cmidrule(lr){2-4}\cmidrule(l){7-8}
    & Rw & Rt & Ra & & & Pixel & First \\
    \midrule
    Swap & \textemdash & \textemdash & \textemdash & 2 & 0.3367 & \textbf{9466.1} & 19512.1 \\
    + Rw & $\checkmark$ & \textemdash & \textemdash & 3 & \textbf{0.0153} & 10652.9 & 11101.8 \\
    + Rw + Ra & $\checkmark$ & \textemdash & $\checkmark$ & 2 & 0.0849 & 9770.2 & \underline{9864.0} \\
    \textbf{Swerve} & $\checkmark$ & $\checkmark$ & $\checkmark$ & 2 & \underline{0.0248} & \underline{9472.3} & \textbf{9605.8} \\
    \bottomrule
  \end{tabular*}
  \endgroup
  \end{minipage}\hfill%
  \begin{minipage}[t]{\columnwidth}
\captionof{table}{\textbf{Swerve with training-free accelerators on HY-WorldPlay.} Accelerators reduce the cost of each chunk and Swerve removes the wait, so the gains add. Speedups are over Wait without acceleration. LI is Light Interaction without its denoising cache.}
  \label{tab:accelerators}
\centering
  \begingroup
  \small
  \renewcommand{\arraystretch}{0.95}
  \setlength{\aboverulesep}{0.25ex}
  \setlength{\belowrulesep}{0.35ex}
  \setlength{\tabcolsep}{2pt}
  \begin{tabular*}{\columnwidth}{@{\extracolsep{\fill}}lcccc@{}}
    \toprule
    & \multicolumn{2}{c}{Throughput$\uparrow$} & \multicolumn{2}{c}{Response speedup$\uparrow$} \\
    \cmidrule(lr){2-3}\cmidrule(l){4-5}
    Accelerator & w/o Swerve & w/ Swerve & w/o Swerve & w/ Swerve \\
    \midrule
    None          & 1.00$\times$ & 1.00$\times$ & 1.00$\times$ & 1.87$\times$ \\
    SageAttention & 1.20$\times$ & 1.20$\times$ & 1.19$\times$ & 2.15$\times$ \\
    LI            & 1.55$\times$ & 1.55$\times$ & 1.40$\times$ & 2.70$\times$ \\
    TeaCache      & 1.11$\times$ & 1.11$\times$ & 1.06$\times$ & 1.87$\times$ \\
    \bottomrule
  \end{tabular*}
  \endgroup
  \end{minipage}
\end{figure*}
\endgroup

\subsection{Ablation and complementarity}
\label{sec:exp-ablation}

\noindent\textbf{Ablation.} \cref{tab:ablation} adds the three operations to Swap one at a time on HY-WorldPlay. Rewrite alone removes 95.5\% of the LPIPS of Swap but adds an evaluation, which delays the first response to 11101.8 ms. Reallocate removes this evaluation, yet the resulting regeneration in two evaluations raises the LPIPS to 0.085, as it lacks the completed refinement. Retain carries this refinement over. Each operation thus covers the cost of the one before it, and together they match the evaluations and pixel-ready latency of Swap at 7\% of its LPIPS. Supplementary material~\ref{sec:supp-ablation} analyzes the sensitivity to $\lambda$ and gives further ablations.

\noindent\textbf{Combination with accelerators.} As shown in \cref{tab:accelerators}, Swerve keeps the number of evaluations of a chunk and leaves how each is computed to the accelerator, so the two do not interfere and each keeps its own gain. Accelerators that reduce the cost of each evaluation also speed up the evaluations after an arrival and thus compound with Swerve, as Light Interaction raises its response speedup from $1.87\times$ to $2.70\times$. TeaCache skips only intermediate evaluations, which Swerve does not run after an arrival, and the two thus keep their own gains without compounding.

\section{Conclusion and limitation}
\label{sec:conclusion}
To conclude, we identify denoising inertia in few-step samplers as the obstacle to in-flight action response in interactive video world models, and propose Swerve, a training-free, plug-and-play denoising-state editor that lets these models respond within the current chunk, down to latent frames. On minWM and HY-WorldPlay, Swerve gives the earliest and most faithful response among in-flight updates. Our evaluation follows the navigation-centered action space of current open world models~\cite{worldplay,genie3}, and the response granularity is bounded by the latent frame at which they take actions. Richer actions and finer action conditioning are promising future directions. Interactive world models still lack a unified metric for real-time response, and we hope our real-time latency helps establish one.

\clearpage
\section*{LLM Usage Statement}
\label{sec:llm-usage}
In this work, we used ChatGPT to polish writing and assist with refining code and experimental procedures. We reviewed all AI-assisted work and take full responsibility for the final content, including text, claims, code, and experimental results.

{
  \small
  \bibliographystyle{ieeenat_fullname}
  \bibliography{main}
}
\input{supplementary}
\end{document}

%% file: supplementary.tex
\clearpage
\setcounter{page}{1}
\renewcommand{\thepage}{S\arabic{page}}
\raggedbottom
\setlength{\stripsep}{9pt plus 2pt minus 2pt}
\pdfbookmark[0]{Appendix}{supplementary-start}
\appendix
\crefalias{section}{appendix}
\setcounter{figure}{0}
\setcounter{table}{0}
\setcounter{equation}{0}
\renewcommand{\thefigure}{S\arabic{figure}}
\renewcommand{\thetable}{S\arabic{table}}
\renewcommand{\theequation}{S\arabic{equation}}
\renewcommand{\theHfigure}{supp.\arabic{figure}}
\renewcommand{\theHtable}{supp.\arabic{table}}
\renewcommand{\theHequation}{supp.\arabic{equation}}

\section*{Appendix}
\noindent This appendix provides additional details and results.
\begin{itemize}[nosep,leftmargin=1em]
\raggedright
  \item the experimental setup, including backbones, baselines, and the evaluation protocol, in~\mbox{\Cref{sec:supp-setup}}.
  \item extended measurements and analysis of denoising inertia and refinement reuse in~\mbox{\Cref{sec:supp-analysis}}.
  \item a sweep over action receipt steps in~\mbox{\Cref{sec:supp-receipt-step}}.
  \item transfer results on additional backbones and broader evaluation of action following in~\mbox{\Cref{sec:supp-backbones}}.
  \item the derivation of response latency under real-time playback in~\mbox{\Cref{sec:supp-realtime}}.
  \item evaluation of sustained interaction (five consecutive action updates) in~\mbox{\Cref{sec:supp-consecutive-updates}}.
  \item sensitivity to the retention coefficient and further ablations in~\mbox{\Cref{sec:supp-ablation}}.
  \item qualitative examples of action responses in~\mbox{\Cref{sec:supp-qualitative}}.
\end{itemize}

\section{Additional experimental setup}
\label{sec:supp-setup}

\noindent\textbf{Backbones.} The two main models generate $832\times480$ video in chunks of four latent frames. For whole-chunk updates, the new action arrives halfway through denoising: $r=2$ for the four-step samplers and $r=4$ for the eight-step sampler. We use $\lambda=\tfrac12$ throughout. \Cref{sec:supp-backbones} details the transfer experiments.
\begin{itemize}[leftmargin=1em,labelsep=0.35em,topsep=2pt,itemsep=1pt,parsep=0pt,partopsep=0pt]
\item \textbf{minWM.} We use minWM-Wan-Action2V-1.3B~\cite{minwm}, conditioned on text and a camera trajectory.
\item \textbf{HY-WorldPlay.} We use the autoregressive 480p I2V distilled model of HY-World1.5 (8B)~\cite{worldplay}, conditioned on text, an initial image, camera poses, and discrete actions.
\item \textbf{ABot-World-0.} We use the 5B-LF checkpoint~\cite{abotworld0} for whole-chunk and frame-accurate updates.
\item \textbf{Yume1.5.} We test whole-chunk updates with its 5B generator and four-step Euler sampler~\cite{yume15}.
\item \textbf{Hunyuan-GameCraft.} To test whether Swerve transfers beyond four-step sampling, we evaluate whole-chunk updates with GameCraft's distilled model using eight denoising steps ($K=8$), with the new action arriving after step four ($r=4$).
\end{itemize}

\smallskip\noindent\textbf{Baselines.}
\begin{itemize}[leftmargin=1em,labelsep=0.35em,topsep=2pt,itemsep=1pt,parsep=0pt,partopsep=0pt]
\item \textbf{Full.} Regenerating the current chunk from the same initial noise under the new action gives the full reference.
\item \textbf{Wait.} The default policy first applies the new action in the next chunk.
\item \textbf{Swap.} Only the condition changes for the remaining evaluations, testing whether they can redirect the chunk.
\item \textbf{Partial rollback.} This diffusion-editing baseline undoes one step and resumes under the new action~\cite{actionsplice,diffusionreroll}.
\item \textbf{Re-noising.} For whole-chunk updates on four-step samplers, we add fresh noise to the latest clean prediction under the old action at the noise level for evaluation 3, then run evaluations 3 and 4 under the new action~\cite{sdedit,actionsplice}. The receipt-step sweep keeps this restart level and two-evaluation budget fixed across $r$.
\item \textbf{ActionSplice\textsuperscript{*}.} Our strongest whole-chunk baseline uses locally trained correctors from the released implementation~\cite{actionsplice}. We use CST-R for whole-chunk updates (\cref{tab:supp-actionsplice-reproduction}) and CST-T for frame-accurate updates (\cref{tab:supp-actionsplice-cst-t-reproduction}).
\end{itemize}

\smallskip\noindent\textbf{Scenes and events.} We use the 54 scenes in \cref{sec:exp-setup}.
\begin{itemize}[leftmargin=1em,labelsep=0.35em,topsep=2pt,itemsep=1pt,parsep=0pt,partopsep=0pt]
\item \textbf{ActionSplice scenes (30).} These include rice terraces, factories, and polar research stations~\cite{actionsplice}.
\item \textbf{Additional scenes (24).} Museum galleries and repair workshops provide close indoor landmarks, while quarry trails and observatory terraces offer long outdoor views.
\end{itemize}
In \cref{tab:main-results}, each scene contributes three switches among forward, backward, and yaw left, and one further event involving an action unseen in corrector training (yaw right, lateral translation, or pitch). All methods share each event's prompt, history, initial noise, and action sequence.

\smallskip\noindent\begin{minipage}{\columnwidth}
\textbf{Interaction settings.}
\begin{itemize}[leftmargin=1em,labelsep=0.35em,topsep=2pt,itemsep=1pt,parsep=0pt,partopsep=0pt]
\item \textbf{Short press.} A temporary action occupies part of a chunk before returning to the previous action.
\item \textbf{Second update.} Two successive in-flight action updates arrive during the same chunk.
\item \textbf{Five consecutive updates.} Five in-flight action updates arrive during one rollout.
\end{itemize}

\noindent We simulate online key presses by revealing each new action to the sampler only at its scheduled arrival during denoising. Until that point, sampling proceeds under the current action.
\end{minipage}\par

\smallskip\noindent\textbf{Evaluation metrics.}\par\nopagebreak[4]
\begin{itemize}[beginpenalty=10000,leftmargin=1em,labelsep=0.35em,topsep=2pt,itemsep=1pt,parsep=0pt,partopsep=0pt]
\item \textbf{Response quality.} LPIPS~\cite{lpips} and PSNR compare decoded frames with Full. \emph{Next} is LPIPS on the next chunk. For a within-chunk change, we score prefix and suffix separately. Boundary error compares the adjacent latent-frame change across the action boundary with the corresponding change in Full. We divide their mean squared difference by the mean squared change in Full, with a denominator floor of $10^{-8}$~\cite{actionsplice}.
\item \textbf{Response latency.} From the key press, \emph{Pixel} measures time until frames generated under the new action are decoded. \emph{First} measures time until the first responding frame plays, using its chunk's delivery time and its playback offset. Following WorldMark~\cite{worldmark}, we detect the response from camera motion estimated by ViPE~\cite{vipe}. A short press counts if both the press and the return are detected, out of all 54 events. Speedup divides the reported First for Wait by the reported First for the method in the same setting. Timing includes native decoding on RTX PRO 6000 Blackwell GPUs.

Real-time latency models generation keeping pace with playback. As in \cref{eq:rt-latency},
\[
T_{\mathrm{RT}}=\Big(1-\frac{r}{K}\Big)L+\frac{n}{K}L+\tau .
\]
At receipt, $(1-r/K)L$ of the preceding chunk remains to play, where $L$ is one chunk's playback duration. The $n$ extra evaluations stall playback by $nL/K$. The video time $\tau$ to the first response is measured from the active chunk's start, including its full duration if the response occurs in the next chunk.
\item \textbf{Visual quality.} VBench~\cite{vbench} is the mean of five scores (subject consistency, background consistency, motion smoothness, aesthetic quality, and imaging quality), reported as a percentage.
\end{itemize}

\par\addvspace{5pt}\noindent\begin{minipage}{\columnwidth}
{
\textbf{ActionSplice reproduction.}
We train the correctors locally following the official ActionSplice procedure~\cite{actionsplice}. To validate our reproduction, we evaluate on the same 30 test scenes and compare LPIPS and PSNR with the reported values for whole-chunk updates (CST-R) and for the edited suffix in single-boundary updates (CST-T). Our results show comparable fidelity to the full reference overall (\cref{tab:supp-actionsplice-reproduction,tab:supp-actionsplice-cst-t-reproduction}).\par
\smallskip
\centering
\begingroup
\renewcommand{\thetable}{A1}
\renewcommand{\theHtable}{supp.actionsplice-reproduction}
\captionsetup{hypcap=false}
\captionof{table}{\textbf{Reproduction of ActionSplice on whole-chunk updates.} The locally trained corrector stays close to the values reported by ActionSplice~\cite{actionsplice} on its 30 test scenes. Reported: values from~\cite{actionsplice}. $*$: trained locally with the official procedure.}
\label{tab:supp-actionsplice-reproduction}
\addtocounter{table}{-1}
\endgroup

\begingroup\small\setlength{\tabcolsep}{4pt}
\begin{tabular*}{\columnwidth}{@{\extracolsep{\fill}}cccc@{}}
\toprule
Backbone & Method & LPIPS $\downarrow$ & PSNR (dB) $\uparrow$ \\
\midrule
\multirow{2}{*}{minWM}
 & CST-R(Reported) & 0.1214 & 16.30 \\
 & CST-R\textsuperscript{*}    & 0.0982 & 18.83 \\
\midrule
\multirow{2}{*}{HY-WorldPlay}
 & CST-R(Reported) & 0.0831 & 20.51 \\
 & CST-R\textsuperscript{*}    & 0.0963 & 20.41 \\
\bottomrule
\end{tabular*}
\endgroup

}\par

\end{minipage}\par\addvspace{5pt}

\par\addvspace{5pt}\noindent\begin{minipage}{\columnwidth}
{
\centering
\begingroup
\renewcommand{\thetable}{A2}
\renewcommand{\theHtable}{supp.actionsplice-cst-t-reproduction}
\captionsetup{hypcap=false}
\captionof{table}{\textbf{Reproduction of ActionSplice on single-boundary updates.} The locally trained CST-T stays close to the reported values on the same scenes. Reported: values from~\cite{actionsplice}. $*$: trained locally with the official procedure.}
\label{tab:supp-actionsplice-cst-t-reproduction}
\addtocounter{table}{-1}
\endgroup

\begingroup\small\setlength{\tabcolsep}{1.7pt}
\begin{tabular*}{\columnwidth}{@{\extracolsep{\fill}}cccc@{}}
\toprule
Backbone & Method & Suffix LPIPS $\downarrow$ & Suffix PSNR $\uparrow$ \\
\midrule
\multirow{2}{*}{HY-WorldPlay}
 & CST-T(Reported) & 0.0615 & 25.05 \\
 & CST-T\textsuperscript{*}    & 0.0627 & 24.89 \\
\bottomrule
\end{tabular*}
\endgroup

}\par

\end{minipage}\par\addvspace{5pt}
\noindent\textbf{Pose-dependent history retrieval.}
In HY-WorldPlay (and other interactive video world models that retrieve history by camera pose), an action update can change which past frames are selected while leaving the committed video history unchanged. In our HY-WorldPlay experiments, all methods use the same native retrieval procedure. Methods that apply the update within the current chunk refresh their context using the revised camera poses, while Wait applies the update at the next chunk.
 
To isolate the role of frame selection, we examine 54 seen-action events, one per scene, for which the native selector returns the entire history. In these events, Swerve and Swap see identical history frames and each uses two post-arrival denoising evaluations. Swerve achieves lower current-chunk LPIPS relative to Full in all 54 paired events (median $0.0259$ for Swerve versus $0.3340$ for Swap). The fidelity advantage therefore persists even when the selected history frames are identical.
\section{Analysis details}
\label{sec:supp-analysis}
\subsection{Denoising inertia across samplers}
\noindent\textbf{Denoising inertia measurement details.}
We evaluate 54 scenes per backbone, with four action-switch events per scene, for a total of 216 events per backbone.

\noindent\textbf{Extended analysis of denoising inertia.}
To further examine the scope of denoising inertia, we extend the analysis from the six four-step samplers in \cref{fig:denoising-inertia} to the eight-step samplers of Astronex and Hunyuan-GameCraft. In \cref{fig:supp-eight-step-inertia}, the measured curves for Astronex and GameCraft both peak at the first step and then gradually decline, further supporting \emph{resistance} in eight-step sampling and showing how it coexists with \emph{persistence}: later denoising steps become progressively less sensitive to new actions, while a perturbation introduced at the first step can persist through to the final output.

\par\addvspace{5pt}\noindent\begin{minipage}{\columnwidth}
\captionsetup{hypcap=false}
\centering
\includegraphics[pagebox=cropbox,width=.98\columnwidth]{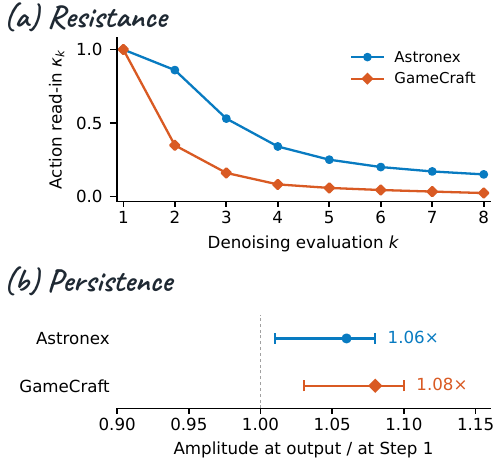}
\captionof{figure}{\textbf{Action read-in and first-step persistence in two eight-step samplers.}}
\label{fig:supp-eight-step-inertia}

\end{minipage}\par\addvspace{5pt}
\subsection{How an early action shapes the final motion}
\label{sec:supp-action-intervention}

Prior work shows that layout and motion emerge early in denoising~\cite{ediffi,prompt2prompt,ac3d,baherwanimotion}. This leaves open a question for interactive generation: once a camera turn has begun to form, can a later action still redirect it? On 30 minWM scenes spanning six action transitions, we hold the history and noise fixed and apply the new action at exactly one of four denoising steps. The other three receive the old action. Two controls use the old or new action throughout.

When only the first step receives the new action, the shift in its clean prediction and the shift in the final latent, both measured relative to the all-old control, have positive cosine similarity in all 30 scenes (median $0.812$). Thus, the first-step perturbation persists in direction even through three subsequent evaluations under the old action. This persistence is also visible in the decoded motion (\cref{fig:supp-single-step-motion}). Across the 21 scenes with measurable camera-motion contrast, changing only the first step recovers a median $90.1\%$ of the signed motion difference between the all-new and all-old controls. Applying the same action only at the second step recovers just $1.7\%$. Together, these results explain why Swerve revisits the first prediction when the user changes direction: the original action's influence persists, while later evaluations have little leverage to redirect the camera. Recomputing the first prediction under the new action recovers the full reference's first prediction in one evaluation.

\par\addvspace{5pt}\noindent\begin{minipage}{\columnwidth}
    \captionsetup{hypcap=false}\centering
    \includegraphics[width=\columnwidth]{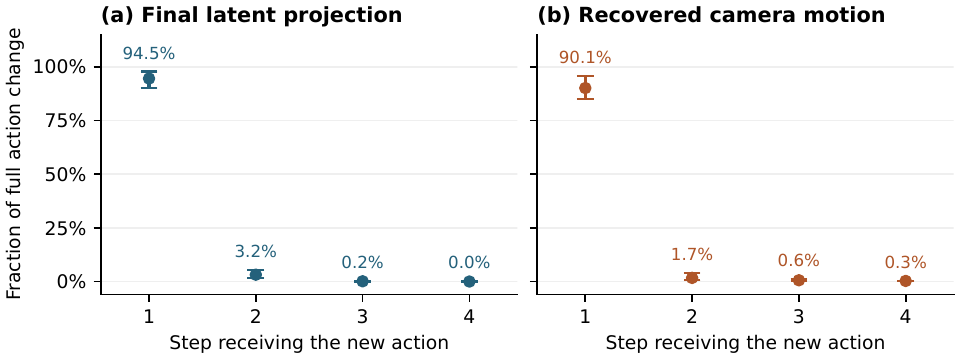}
    \captionof{figure}{\textbf{Single-step action intervention on minWM.}
    The horizontal axis indicates the only denoising step that receives the new action.
    (a) Final latent shift projected onto the all-new minus all-old latent difference and normalized by its magnitude (30 scenes).
    (b) Fraction of the corresponding signed change in camera yaw or longitudinal displacement recovered in the decoded video.
    Markers show medians and whiskers show interquartile ranges.}
    \label{fig:supp-single-step-motion}

\end{minipage}\par\addvspace{5pt}
\subsection{Reusable refinement}
Retain scales the old refinement to adjust its magnitude while preserving its direction. The more closely the old and new refinements align, the greater the attainable reduction in anchor error. Across the eight backbones in \cref{tab:supp-refinement}, the representative retention coefficients range from 0.36 to 0.63, all within 0.14 of 0.5. Fixing $\lambda=0.5$ reduces representative anchor error in all eight settings, by 14.0\%--74.7\%. Thus, for the backbone settings in \cref{tab:supp-refinement}, $\lambda=0.5$ can serve as a fixed hyperparameter across backbones, without per-backbone tuning. To obtain a representative coefficient more closely tailored to a particular backbone, we run the old and new actions separately on development inputs up to the receipt step to obtain paired predictions. The median cosine similarity $\bar c$ and median norm ratio $\bar s$ then give $\lambda_{\mathrm{rep}}=\operatorname{clip}(\bar c/\bar s,0,1)$ directly. This calibration requires no training, decoding, or parameter sweep.

\par\addvspace{5pt}\noindent\begin{minipage}{\columnwidth}
\captionsetup{hypcap=false}
\centering
\captionof{table}{\textbf{Reusable refinement at halfway receipt.} The representative retention coefficient $\lambda_{\mathrm{rep}}$ minimizes the anchor error for the backbone's median cosine similarity $\bar c$ and median old-to-new norm ratio $\bar s$. Gain denotes the percentage reduction in anchor error at $\lambda=0.5$ relative to retaining no refinement ($\lambda=0$).}
\label{tab:supp-refinement}
\begingroup\small\setlength{\tabcolsep}{2.5pt}
\begin{tabular*}{\linewidth}{@{\extracolsep{\fill}}cccccc@{}}
\toprule
Backbone & $r$ & $\bar c$ & $\bar s$ & $\lambda_{\mathrm{rep}}$ & Gain (\%) \\
\midrule
minWM & 2 & 0.90 & 1.60 & 0.56 & 55.3 \\
HY-WorldPlay & 2 & 0.97 & 1.80 & 0.54 & 74.7 \\
ABot-World-0 & 2 & 0.88 & 1.77 & 0.50 & 52.5 \\
Matrix-Game & 2 & 0.91 & 1.48 & 0.61 & 55.2 \\
LingBot-World & 2 & 0.65 & 1.37 & 0.47 & 23.9 \\
Yume1.5 & 2 & 0.55 & 1.51 & 0.36 & 14.0 \\
Astronex & 4 & 0.87 & 1.39 & 0.63 & 47.7 \\
Hunyuan-GameCraft & 4 & 0.57 & 1.50 & 0.38 & 15.9 \\
\bottomrule
\end{tabular*}
\endgroup

\end{minipage}\par\addvspace{5pt}
\section{Receipt step sweep}
\label{sec:supp-receipt-step}

\Cref{tab:supp-receipt-sweep} varies the receipt step on the 54 scenes of \cref{tab:main-results}. At $r=1$, no refinement has been completed under the old action, so Swerve reduces to the rewrite followed by the two remaining evaluations and reaches its lowest LPIPS, 0.023 on minWM and 0.014 on HY-WorldPlay. Its real-time latency is longer than at $r=2$, as more of the preceding chunk remains to play when the key is pressed.

At $r=3$, the single remaining evaluation cannot be reassigned, so Swerve runs the rewrite and the final evaluation, one more than ordinary sampling. Returning the rewritten anchor without the final evaluation saves this evaluation and responds sooner, at a higher LPIPS of 0.080 and 0.054 against 0.037 and 0.025.

At both receipt steps, Swerve responds earlier and more faithfully than ActionSplice\textsuperscript{*}. The corrector degrades when it acts late, and its LPIPS at $r=3$ rises to the level of Wait, consistent with the receipt step sweep of ActionSplice~\cite{actionsplice}. On HY-WorldPlay, part of its events respond only in the next chunk, so its median first response falls between the two chunks.
\section{Further backbones and broader evaluations}
\label{sec:supp-backbones}

\noindent\textbf{Further Backbones.}
\Cref{tab:supp-transfer-whole,tab:supp-transfer-frame} apply Swerve to ABot-World-0, Yume1.5 and GameCraft with $\lambda=\tfrac12$ and no other change. In whole-chunk updates, Swerve responds $2.67\times$, $2.29\times$ and $3.17\times$ earlier than Wait.

In frame-accurate updates on ABot-World-0, Swerve detects 53 of the 54 short presses against none for the baselines and responds first to a second update.

\noindent\textbf{Broader Evaluations.}
Alongside the full-reference fidelity metrics in \cref{tab:main-results}, we evaluate how accurately and promptly the generated video follows the revised action. \Cref{tab:supp-camera-action} compares camera trajectories estimated by ViPE with the trajectory specified by the input commands and complements this comparison with two human judgments of the visible response.

\makeatletter
\newcommand{\SuppFullWidthRow}{%
  \noalign{\vbox to 0pt{\color{black!10}%
    \hrule width \textwidth
           height \dimexpr\ht\@arstrutbox+\dp\@arstrutbox\relax
    \vss}}}
\makeatother

\clearpage
\begin{table*}[p]
\noindent\begin{minipage}{\textwidth}
{
\centering\captionsetup{hypcap=false}
\captionof{table}{\textbf{Receipt step sweep on minWM and HY-WorldPlay.} Swerve responds earlier and more faithfully than ActionSplice\textsuperscript{*} at both receipt steps. Events follow \cref{tab:main-results} on the 54 scenes, with the new action arriving after $r=1$ or $3$ of $K=4$ evaluations. Evals: evaluations of the current chunk after the arrival. Pixel, First: end-to-end pixel-ready, first-response latency in ms. RT: real-time latency, \cref{eq:rt-latency}. Speedup: over Wait at the same $r$. $\dagger$: reused from \cref{tab:main-results}, for Wait in response quality only.}
\label{tab:supp-receipt-sweep}
\begingroup\footnotesize\setlength{\tabcolsep}{6.4pt}\renewcommand{\arraystretch}{1.02}
\makeatletter
\newcommand{\SuppFullRow}{%
  \noalign{\vbox to 0pt{\color{black!10}%
    \setbox\z@=\hbox{Full$^{\dagger}$}%
    \dimen@=\ht\@arstrutbox
    \ifdim\ht\z@>\dimen@ \dimen@=\ht\z@\fi
    \advance\dimen@ by \dp\@arstrutbox\relax
    \hrule width \textwidth height \dimen@
    \vss}}}
\makeatother
\begin{tabular*}{\textwidth}{@{\extracolsep{\fill}}ccccccccccc@{}}
\noalign{\vskip 5pt\hbox{\strut\bfseries (a) minWM}\vskip 2pt}
\toprule
 & & & \multicolumn{4}{c}{Latency} & \multicolumn{3}{c}{Response quality} & Visual quality \\
\cmidrule(lr){4-7}\cmidrule(lr){8-10}\cmidrule(l){11-11}
$r$ & Method & Evals. & Pixel$\downarrow$ & First$\downarrow$ & RT$\downarrow$ & Speedup$\uparrow$ & LPIPS$\downarrow$ & PSNR$\uparrow$ & Next$\downarrow$ & VBench (\%)$\uparrow$ \\
\midrule
\SuppFullRow
\multirow{4}{*}{\textbf{1}} & Full$^{\dagger}$ & \underline{4} & 2103.0 & 3291.8 & 1250.0 & 1.37$\times$ & --- & --- & --- & 86.02 \\
 & Wait$^{\dagger}$ & \textbf{3} & 4205.1 & 4513.5 & 2062.5 & 1.00$\times$ & 0.1292 & 17.48 & 0.3396 & 86.11 \\
 & ActionSplice\textsuperscript{*} & \textbf{3} & \underline{1726.5} & \underline{1994.3} & \underline{1062.5} & \underline{2.26$\times$} & \underline{0.0616} & \underline{20.08} & \underline{0.1960} & 85.78 \\
 & \textbf{Swerve} & \textbf{3} & \textbf{1709.6} & \textbf{1976.0} & \textbf{1000.0} & \textbf{2.28$\times$} & \textbf{0.0231} & \textbf{25.67} & \textbf{0.0602} & 85.86 \\
\midrule
\SuppFullRow
\multirow{5}{*}{\textbf{3}} & Full$^{\dagger}$ & 4 & 2103.0 & 3291.8 & 1250.0 & 1.15$\times$ & --- & --- & --- & 86.02 \\
 & Wait$^{\dagger}$ & \textbf{1} & 3463.2 & 3777.1 & 1562.5 & 1.00$\times$ & 0.1292 & 17.48 & 0.3396 & 86.11 \\
 & ActionSplice\textsuperscript{*} & \textbf{1} & \underline{985.4} & 3570.6 & 1484.4 & 1.06$\times$ & 0.1329 & 18.49 & 0.3187 & 85.36 \\
 & \textbf{Swerve} & \underline{2} & 1348.9 & \underline{1592.5} & \underline{750.0} & \underline{2.37$\times$} & \textbf{0.0366} & \textbf{23.23} & \textbf{0.0885} & 85.88 \\
 & Swerve (anchor only) & \textbf{1} & \textbf{981.1} & \textbf{1234.7} & \textbf{500.0} & \textbf{3.06$\times$} & \underline{0.0798} & \underline{21.30} & \underline{0.1249} & 85.82 \\
\bottomrule
\noalign{\vskip 10pt\hbox{\strut\bfseries (b) HY-WorldPlay}\vskip 2pt}
\toprule
 & & & \multicolumn{4}{c}{Latency} & \multicolumn{3}{c}{Response quality} & Visual quality \\
\cmidrule(lr){4-7}\cmidrule(lr){8-10}\cmidrule(l){11-11}
$r$ & Method & Evals. & Pixel$\downarrow$ & First$\downarrow$ & RT$\downarrow$ & Speedup$\uparrow$ & LPIPS$\downarrow$ & PSNR$\uparrow$ & Next$\downarrow$ & VBench (\%)$\uparrow$ \\
\midrule
\SuppFullRow
\multirow{4}{*}{\textbf{1}} & Full$^{\dagger}$ & \underline{4} & 13696.8 & 13819.6 & 750.0 & 1.40$\times$ & --- & --- & --- & 86.31 \\
 & Wait$^{\dagger}$ & \textbf{3} & 19149.8 & 19293.2 & 1291.7 & 1.00$\times$ & 0.3395 & 13.56 & 0.5184 & 86.17 \\
 & ActionSplice\textsuperscript{*} & \textbf{3} & \textbf{10778.7} & \underline{11000.4} & \underline{666.7} & \underline{1.75$\times$} & \underline{0.1259} & \underline{17.12} & \underline{0.1877} & 85.38 \\
 & \textbf{Swerve} & \textbf{3} & \underline{10782.1} & \textbf{10901.8} & \textbf{593.8} & \textbf{1.77$\times$} & \textbf{0.0141} & \textbf{30.29} & \textbf{0.0406} & 86.15 \\
\midrule
\SuppFullRow
\multirow{5}{*}{\textbf{3}} & Full$^{\dagger}$ & 4 & 13696.8 & 13819.6 & 750.0 & 1.22$\times$ & --- & --- & --- & 86.31 \\
 & Wait$^{\dagger}$ & \textbf{1} & 16786.9 & 16920.2 & 958.3 & 1.00$\times$ & 0.3395 & 13.56 & 0.5184 & 86.17 \\
 & ActionSplice\textsuperscript{*} & \textbf{1} & \textbf{8466.2} & 14482.5 & 822.9 & 1.17$\times$ & 0.3303 & 15.23 & 0.4771 & 83.63 \\
 & \textbf{Swerve} & \underline{2} & 9853.5 & \underline{9968.5} & \underline{416.7} & \underline{1.70$\times$} & \textbf{0.0249} & \textbf{27.39} & \textbf{0.0579} & 86.13 \\
 & Swerve (anchor only) & \textbf{1} & \underline{8723.2} & \textbf{8848.2} & \textbf{250.0} & \textbf{1.91$\times$} & \underline{0.0539} & \underline{24.29} & \underline{0.0929} & 86.10 \\
\bottomrule
\end{tabular*}\endgroup

}\par
\par\vspace{12pt}
{
\centering\captionsetup{hypcap=false}
\captionof{table}{\textbf{Whole-chunk transfer to ABot-World-0, Yume1.5 and GameCraft.} Swerve responds earliest and most faithfully on all three backbones with $\lambda=\tfrac12$ unchanged. Events follow \cref{tab:main-results} on the 54 scenes, with $K=4$ and $r=2$ for ABot-World-0 and Yume1.5, and $K=8$ and $r=4$ for GameCraft. Columns follow \cref{tab:main-results}.}
\label{tab:supp-transfer-whole}
\begingroup\footnotesize\setlength{\tabcolsep}{2.4pt}
\renewcommand{\arraystretch}{1.02}
\begin{tabular*}{\textwidth}{@{\extracolsep{\fill}}cccccccccc@{}}
\toprule
 & & \multicolumn{4}{c}{Latency} & \multicolumn{3}{c}{Response quality} & Visual quality \\
\cmidrule(lr){3-6}\cmidrule(lr){7-9}\cmidrule(l){10-10}
Backbone & Method & Pixel$\downarrow$ & First$\downarrow$ & RT$\downarrow$ & Speedup$\uparrow$ &
LPIPS$\downarrow$ & PSNR$\uparrow$ & Next$\downarrow$ & VBench (\%)$\uparrow$ \\
\midrule
\SuppFullWidthRow
\multirow{6}{*}{\textbf{ABot}} & Full & 896.5 & \underline{1004.1} & \underline{1083.3} & \underline{1.75$\times$} & \textemdash & \textemdash & \textemdash & 84.15 \\
 & Wait & 1524.2 & 1761.4 & 1750.0 & 1.00$\times$ & 0.3519 & 18.03 & 0.4732 & 84.17 \\
 & Swap & \underline{562.8} & 1771.9 & 1750.0 & 0.99$\times$ & 0.3489 & 18.01 & 0.4808 & 84.29 \\
 & Partial rollback & 732.2 & 1927.3 & 2000.0 & 0.91$\times$ & \underline{0.3433} & \underline{18.17} & \underline{0.4693} & 84.17 \\
 & Re-noising & \textbf{562.4} & 1806.4 & 1750.0 & 0.98$\times$ & 0.3500 & 17.91 & 0.4808 & 84.03 \\
 & \textbf{Swerve} & \textbf{562.4} & \textbf{658.9} & \textbf{583.3} & \textbf{2.67$\times$} & \textbf{0.0415} & \textbf{23.92} & \textbf{0.0865} & 84.37 \\
\midrule
\SuppFullWidthRow
\multirow{6}{*}{\textbf{Yume 1.5}} & Full & 10815.2 & \underline{11025.5} & \underline{2000.0} & \underline{1.94$\times$} & \textemdash & \textemdash & \textemdash & 82.97 \\
 & Wait & 20712.9 & 21434.7 & 3406.2 & 1.00$\times$ & 0.3922 & \underline{14.84} & 0.5939 & 82.54 \\
 & Swap & \underline{9143.3} & 21440.6 & 3406.2 & 1.00$\times$ & 0.3917 & \underline{14.84} & \underline{0.5935} & 82.48 \\
 & Partial rollback & 9985.0 & 22276.1 & 3828.1 & 0.96$\times$ & \underline{0.3912} & 14.83 & \underline{0.5935} & 82.54 \\
 & Re-noising & 9145.5 & 21328.7 & 3406.2 & 1.00$\times$ & 0.3976 & 14.80 & 0.5971 & 82.50 \\
 & \textbf{Swerve} & \textbf{9142.6} & \textbf{9344.9} & \textbf{1093.8} & \textbf{2.29$\times$} & \textbf{0.0618} & \textbf{23.49} & \textbf{0.0951} & 82.20 \\
\midrule
\SuppFullWidthRow
\multirow{6}{*}{\textbf{GameCraft}}
 & Full & 209819.4 & \underline{220411.8} & \underline{1416.7} & \underline{1.60$\times$}
 & \textemdash & \textemdash & \textemdash & 84.41 \\
 & Wait & 332303.6 & 352999.0 & 2104.2 & 1.00$\times$
 & \underline{0.5208} & \underline{12.26} & 0.6161 & 83.88 \\
 & Swap & \textbf{110602.5} & 324513.0 & 2104.2 & 1.09$\times$
 & 0.5213 & \underline{12.26} & 0.6163 & 83.79 \\
 & Partial rollback & 135192.5 & 349871.7 & 2276.0 & 1.01$\times$
 & 0.5215 & \underline{12.26} & 0.6163 & 83.78 \\
 & Re-noising & 111326.8 & 347974.0 & 2104.2 & 1.01$\times$
 & 0.5267 & 12.22 & \underline{0.6121} & 83.97 \\
 & \textbf{Swerve} & \underline{110983.0} & \textbf{111196.9} & \textbf{729.2} & \textbf{3.17$\times$}
 & \textbf{0.0398} & \textbf{26.43} & \textbf{0.0695} & 83.81 \\
\bottomrule
\end{tabular*}\endgroup
\par

}\par

\end{minipage}\par
\end{table*}

\clearpage
\begin{strip}
\noindent\begin{minipage}{\textwidth}
{
\centering\captionsetup{hypcap=false}
\captionof{table}{\textbf{Frame-accurate transfer to ABot-World-0.} Swerve leads on the suffix and responds earliest in every setting. Settings and columns follow \cref{tab:frame-response} on the 54 scenes.}
\label{tab:supp-transfer-frame}
\begingroup\small\setlength{\tabcolsep}{2.2pt}
\renewcommand{\arraystretch}{1.24}
\begin{tabular*}{\textwidth}{@{\extracolsep{\fill}}cccccccccccc@{}}
\toprule
 & \multicolumn{5}{c}{Single boundary} & \multicolumn{3}{c}{Short press} & \multicolumn{3}{c}{Second update} \\
\cmidrule(lr){2-6}\cmidrule(lr){7-9}\cmidrule(l){10-12}
Method & \shortstack{Prefix\\LPIPS$\downarrow$} & \shortstack{Suffix\\LPIPS$\downarrow$} &
\shortstack{Suffix\\PSNR$\uparrow$} & \shortstack{Boundary\\error$\downarrow$} & First$\downarrow$ &
\shortstack{Suffix\\LPIPS$\downarrow$} & \shortstack{Suffix\\PSNR$\uparrow$} & \shortstack{Detected\\events$\uparrow$} &
\shortstack{Suffix\\LPIPS$\downarrow$} & \shortstack{Suffix\\PSNR$\uparrow$} & First$\downarrow$ \\
\midrule
\SuppFullWidthRow
Full & \textemdash & \textemdash & \textemdash & \textemdash & 4275.6 & \textemdash & \textemdash & 54/54 & \textemdash & \textemdash & 3972.6 \\
Swap & 0.0278 & 0.3991 & 18.08 & 0.8489 & 4621.9 & 0.3365 & 18.93 & 0/54 & 0.3928 & 16.87 & 3166.6 \\
Partial rollback & 0.0630 & \underline{0.3745} & \underline{18.61} & \underline{0.8414} & 4599.4 & \underline{0.3270} & \underline{19.20} & 0/54 & \underline{0.3230} & \underline{19.31} & 3312.5 \\
Re-noising & 0.0542 & 0.4041 & 17.76 & 0.8906 & \underline{4550.2} & 0.3403 & 18.71 & 0/54 & 0.4071 & 16.89 & \underline{3195.5} \\
\textbf{Swerve} & 0.0749 & \textbf{0.0674} & \textbf{24.74} & \textbf{0.1006} & \textbf{2127.3} & \textbf{0.0610} & \textbf{25.27} & \textbf{53/54} & \textbf{0.0849} & \textbf{25.04} & \textbf{2273.4} \\
\bottomrule
\end{tabular*}\endgroup
\par

}\par

\end{minipage}\par
\end{strip}
\begin{itemize}[leftmargin=1em,labelsep=0.35em,topsep=2pt,itemsep=1pt,parsep=0pt,partopsep=0pt]
\item \textbf{Rotation error ($R_{\mathrm{dist}}$).}
We align the estimated and commanded trajectories to their respective first available poses and report the mean angular difference between camera orientations over the updated chunk, in radians.
\item \textbf{Translation error ($T_{\mathrm{dist}}$).}
To account for monocular scale ambiguity, we rescale the estimated translations to match the commanded displacement magnitude at the time of maximum commanded displacement. We then report the mean squared distance between the estimated and commanded camera positions over the updated chunk.

\item \textbf{Action following.}
Annotators judge whether the revised action becomes visible within the editable region. We report the percentage of evaluated events for which they observe the requested change.

\item \textbf{Mean stale frames.}
Annotators identify the first frame that visibly responds to the revised action. We count the frames from the start of the editable region up to, but excluding, that frame and report the mean count. A response in the first editable frame therefore incurs zero stale frames.
\end{itemize}

\par\addvspace{5pt}\noindent\begin{minipage}{\columnwidth}
\captionsetup{hypcap=false}\centering
\captionof{table}{\textbf{Camera-action following under whole-chunk updates.}}
\label{tab:supp-camera-action}
\begingroup
\small
\setlength{\tabcolsep}{1.5pt}
\renewcommand{\arraystretch}{1.08}

\makeatletter
\newcommand{\CameraFullRow}{%
  \noalign{\vbox to 0pt{\color{black!10}%
    \hrule width \columnwidth
           height \dimexpr\ht\@arstrutbox+\dp\@arstrutbox\relax
    \vss}}}
\makeatother

\begin{tabular*}{\columnwidth}{@{\extracolsep{\fill}}cccccc@{}}
\toprule
 & & \multicolumn{2}{c}{ViPE} & \multicolumn{2}{c}{Human} \\
\cmidrule(lr){3-4}\cmidrule(l){5-6}
Backbone & Method
 & $R_{\mathrm{dist}}{\downarrow}$
 & $T_{\mathrm{dist}}{\downarrow}$
 & \shortstack{Action\\following (\%)$\uparrow$}
 & \shortstack{Mean stale\\frames$\downarrow$} \\
\midrule
\CameraFullRow
\multirow{3}{*}{minWM}
 & Full            & 0.0961          & 0.1742          & \textbf{100} & \textbf{0.0} \\
 & Wait            & 0.0984          & 0.1907          & 6.2 & 15.8\\
 & \textbf{Swerve} & \textbf{0.0937} & \textbf{0.1729} & \textbf{100} & \underline{0.7}\\
\midrule
\CameraFullRow
\multirow{3}{*}{\shortstack{HY-\\WorldPlay}}
 & Full            & 0.0537          & 0.0131          & \textbf{100} & \textbf{0.0} \\
 & Wait            & 0.0699          & 0.0729          & 5.6 & 15.9\\
 & \textbf{Swerve} & \textbf{0.0466} & \textbf{0.0113} & \textbf{100} & \underline{0.2} \\
\bottomrule
\end{tabular*}
\endgroup

\end{minipage}\par\addvspace{5pt}
\newpage
\section{Response under real-time generation}
\label{sec:supp-realtime}

\noindent\textbf{Operating point.}
Interactive world models are deployed as streaming pipelines, in which the next chunk is generated while the current one plays and frames are decoded progressively for display~\cite{worldplay,minwm}. HY-WorldPlay is served in this way at 24~FPS on multiple GPUs~\cite{worldplay}, so that generation keeps pace with playback. The real-time latency evaluates the response at this operating point, where the $K$ evaluations of a chunk span the playback time $L$ of the preceding chunk and each evaluation takes $L/K$.

\smallskip
\noindent\textbf{Derivation.}
A key press is received by the sampler between two evaluations. At receipt step $r$, the chunk being denoised is the current chunk, and the preceding chunk is on screen with $(1-r/K)L$ of it left to play. An update that keeps the number of evaluations delivers the current chunk exactly when the preceding one ends, so playback continues without interruption. Each evaluation that an update adds delays this delivery by $L/K$ and holds the last displayed frame for that time, which gives $nL/K$. An update that runs fewer evaluations still waits for the preceding chunk to end, so $n$ is never negative. Once the current chunk plays, the first frame that follows the new action appears after the video time $\tau$, which covers the whole current chunk if the response first occurs in the next one. The sum of the three parts is given by \cref{eq:rt-latency}.

\clearpage
\begin{strip}
\noindent\begin{minipage}{\textwidth}
\begin{minipage}[t]{\dimexpr(\textwidth-\columnsep)/2\relax}\vspace{0pt}
\smallskip
\begin{samepage}
\noindent\textbf{Interpretation.}
Each term of \cref{eq:rt-latency} is fixed by the pipeline, counted from the evaluations of an update, or measured on its generated frames, and none depends on the GPU that produced the frames. The real-time latency therefore represents the delay that a user perceives between a key press and the response on screen in a deployment that keeps pace with playback. The end-to-end latency follows the same pipeline at the speed of a single GPU, which is slower than playback, and shares the video time $\tau$ with the real-time latency. In \cref{tab:main-results}, Swerve achieves the lowest First and RT latencies on both backbones.

\end{samepage}
\end{minipage}\hfill
\begin{minipage}[t]{\dimexpr(\textwidth-\columnsep)/2\relax}\vspace{0pt}
\section{Consecutive action updates}
\label{sec:supp-consecutive-updates}

\Cref{tab:supp-five-updates} applies five updates in succession to each of the 54 scenes. The LPIPS of Swerve grows with the number of updates, from 0.040 to 0.385 on minWM and from 0.023 to 0.277 on HY-WorldPlay, as each chunk enters the history of the next and the rollout drifts from the full reference. It stays below that of Swap and ActionSplice\textsuperscript{*} after every update, and all five responses of Swerve are detected in 52 and 51 of the 54 scenes, against at most 7 for the baselines.
\end{minipage}
\par
\par\vspace{14pt}
{
\centering\captionsetup{hypcap=false}
\captionof{table}{\textbf{Five consecutive action updates on minWM and HY-WorldPlay.} The error of Swerve grows with each update but stays below that of the baselines throughout. Each of the 54 scenes receives five updates in succession, each after $r=2$ of $K=4$ evaluations, with action sequences sampled separately from those of \cref{tab:main-results}. 1--5: LPIPS of the current chunk after each update. Detected events: scenes in which all five responses are detected.}
\label{tab:supp-five-updates}
\begingroup\small\setlength{\tabcolsep}{2.8pt}
\renewcommand{\arraystretch}{1.48}
\begin{tabular*}{\textwidth}{@{\extracolsep{\fill}}ccccccccc@{}}
\toprule
 & & \multicolumn{5}{c}{Active-chunk LPIPS $\downarrow$} & \multicolumn{1}{c}{Response detection} & Visual quality \\
\cmidrule(lr){3-7}\cmidrule(lr){8-8}\cmidrule(l){9-9}
Backbone & Method & 1 & 2 & 3 & 4 & 5 & Detected events$\uparrow$ & VBench (\%) $\uparrow$ \\
\midrule
\SuppFullWidthRow
\multirow{4}{*}{\textbf{minWM}} & Full & \textemdash & \textemdash & \textemdash & \textemdash & \textemdash & 54/54 & 83.45 \\
 & Swap & 0.2165 & 0.5386 & 0.5550 & 0.5669 & 0.5741 & 0/54 & 83.50 \\
 & ActionSplice\textsuperscript{*} & \underline{0.1947} & \underline{0.5307} & \underline{0.5441} & \underline{0.5513} & \underline{0.5649} & \underline{2/54} & 83.32 \\
 & \textbf{Swerve} & \textbf{0.0401} & \textbf{0.1457} & \textbf{0.2231} & \textbf{0.3292} & \textbf{0.3845} & \textbf{52/54} & 83.39 \\
\midrule
\SuppFullWidthRow
\multirow{4}{*}{\textbf{HY-WorldPlay}} & Full & \textemdash & \textemdash & \textemdash & \textemdash & \textemdash & 54/54 & 83.26 \\
 & Swap & 0.3569 & 0.4437 & \underline{0.4168} & \underline{0.4461} & \underline{0.4612} & 0/54 & 82.93 \\
 & ActionSplice\textsuperscript{*} & \underline{0.2829} & \underline{0.4163} & 0.4300 & 0.4848 & 0.5408 & \underline{7/54} & 81.73 \\
 & \textbf{Swerve} & \textbf{0.0226} & \textbf{0.1128} & \textbf{0.1483} & \textbf{0.2120} & \textbf{0.2766} & \textbf{51/54} & 83.28 \\
\bottomrule
\end{tabular*}\endgroup

}\par

\end{minipage}\par
\end{strip}
\begin{strip}
\noindent\begin{minipage}{\textwidth}
\begin{minipage}[t]{\dimexpr(\textwidth-\columnsep)/2\relax}\vspace{0pt}
\section{Hyperparameter sensitivity and further ablations}
\label{sec:supp-ablation}
\noindent\textbf{Retention coefficient.} The coefficient $\lambda$ controls how much of the old refinement is retained. We analyze it per backbone because the alignment and relative magnitude of the old and new refinements vary across backbones. \Cref{sec:supp-analysis} defines the representative coefficient and describes how to calibrate it.
\end{minipage}\hfill
\begin{minipage}[t]{\dimexpr(\textwidth-\columnsep)/2\relax}\vspace{0pt}
\centering\captionsetup{hypcap=false}
\captionof{table}{\textbf{Extended ablation of the three operations on HY-WorldPlay.} Retain lowers the error with or without Reallocate. Rw, Rt, Ra: Rewrite, Retain, Reallocate. Evals: evaluations after the arrival. Pixel, First: pixel-ready, first-response latency in ms. All rows except + Rw + Rt reproduce \cref{tab:ablation}.}
\label{tab:supp-extended-ablation}
\begingroup\small\setlength{\tabcolsep}{1pt}
\renewcommand{\arraystretch}{1.35}
\begin{tabular*}{\linewidth}{@{\extracolsep{\fill}}cccccccc@{}}
    \toprule
    \multirow{2}{*}{Variant} & \multicolumn{3}{c}{Components} & \multirow{2}{*}{Evals.} & \multirow{2}{*}{LPIPS$\downarrow$} & \multicolumn{2}{c}{Latency (ms)$\downarrow$} \\
    \cmidrule(lr){2-4}\cmidrule(l){7-8}
    & Rw & Rt & Ra & & & Pixel & First \\
    \midrule
    Swap & \textemdash & \textemdash & \textemdash & 2 & 0.3367 & 9466.1 & 19512.1 \\
    + Rw & $\checkmark$ & \textemdash & \textemdash & 3 & 0.0153 & 10652.9 & 11101.8 \\
    + Rw + Rt & $\checkmark$ & $\checkmark$ & \textemdash & 3 & 0.0141 & 10641.8 & 11103.7 \\
    + Rw + Ra & $\checkmark$ & \textemdash & $\checkmark$ & 2 & 0.0849 & 9770.2 & 9864.0 \\
    \textbf{Swerve} & $\checkmark$ & $\checkmark$ & $\checkmark$ & 2 & 0.0248 & 9472.3 & 9605.8 \\
    \bottomrule
  \end{tabular*}
\endgroup
\end{minipage}
\par

\end{minipage}\par
\end{strip}
\clearpage

\begingroup
\begin{strip}
\captionsetup{hypcap=false}
\noindent\begin{minipage}{\dimexpr(\textwidth-\columnsep)/2\relax}
\section{Representative Visualizations}
\label{sec:supp-qualitative}

We compare Swap and Swerve on short presses and whole-chunk action
updates. In the whole-chunk examples, the first column shows the
shared history. Times are measured on the video timeline from the update.
\end{minipage}\par\medskip

\par\smallskip\noindent
\begin{minipage}{\linewidth}
  \centering
  \captionsetup{hypcap=false,skip=4pt}
  \includegraphics[width=\linewidth]{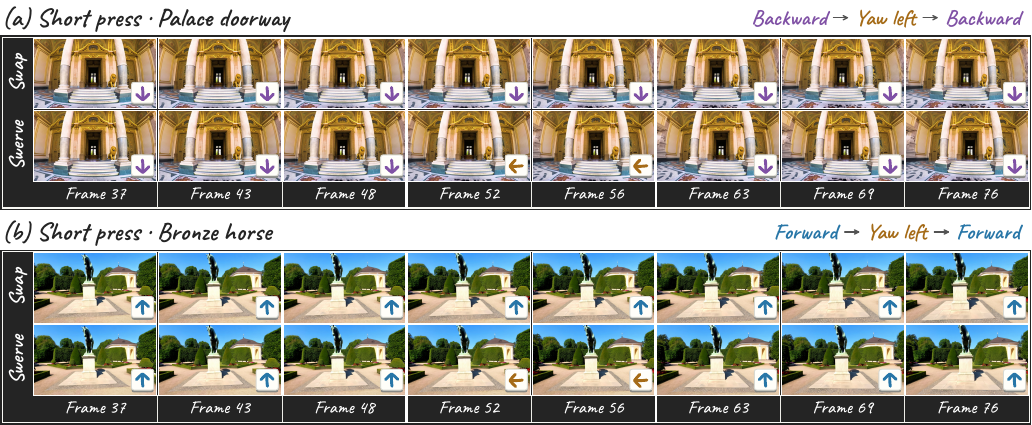}%
  \captionof{figure}{\textbf{Short presses at the palace doorway and the bronze horse.}
   The middle two sampled frames lie within the brief yaw-left request.
   The last three follow the return to the original action.}%
  \label{fig:supp-short-press-palace-horse}%
\end{minipage}\par

\medskip
\par\smallskip\noindent
\begin{minipage}{\linewidth}
  \centering
  \captionsetup{hypcap=false,skip=4pt}
  \includegraphics[width=\linewidth]{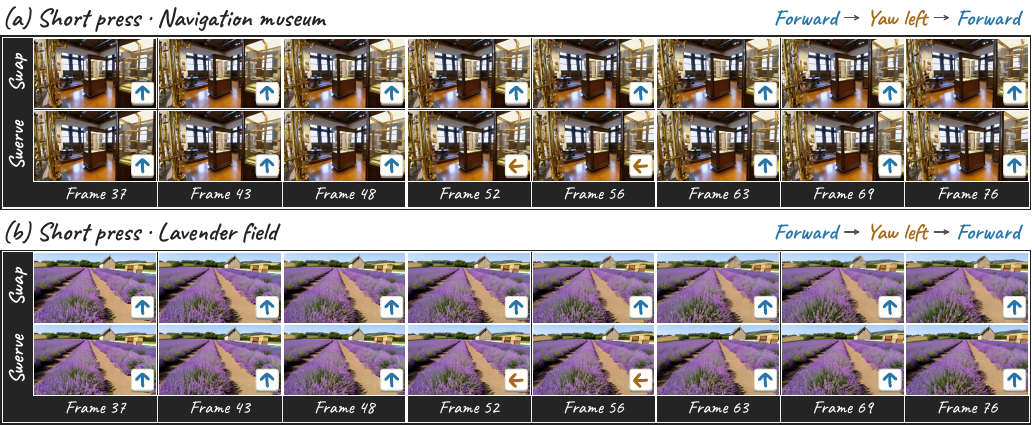}%
  \captionof{figure}{\textbf{Short presses at the navigation museum and the lavender field.}
   The middle two sampled frames lie within the brief yaw-left request.
   The last three follow the return to forward motion.}%
  \label{fig:supp-short-press-museum-lavender}%
\end{minipage}\par
\end{strip}

\clearpage

\begin{figure*}[p]
\par\smallskip\noindent
\begin{minipage}{\linewidth}
  \centering
  \captionsetup{hypcap=false,skip=4pt}
  \includegraphics[width=.87\linewidth]{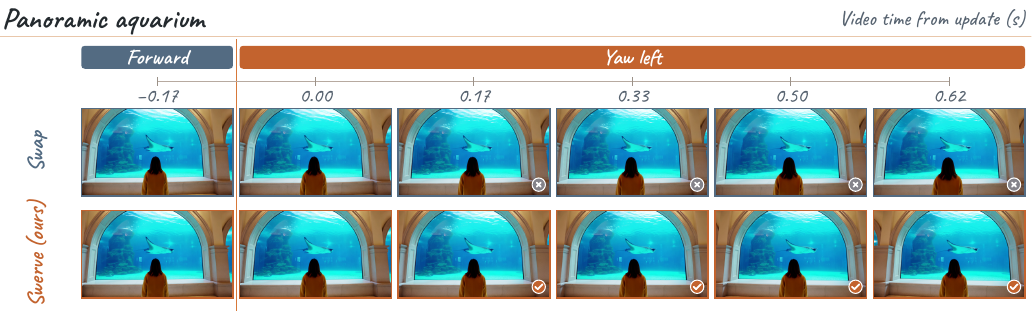}%
  \captionof{figure}{\textbf{Panoramic aquarium.} The action switches from forward to yaw left.}%
  \label{fig:supp-whole-aquarium}%
\end{minipage}\par

\par\smallskip\noindent
\begin{minipage}{\linewidth}
  \centering
  \captionsetup{hypcap=false,skip=4pt}
  \includegraphics[width=.87\linewidth]{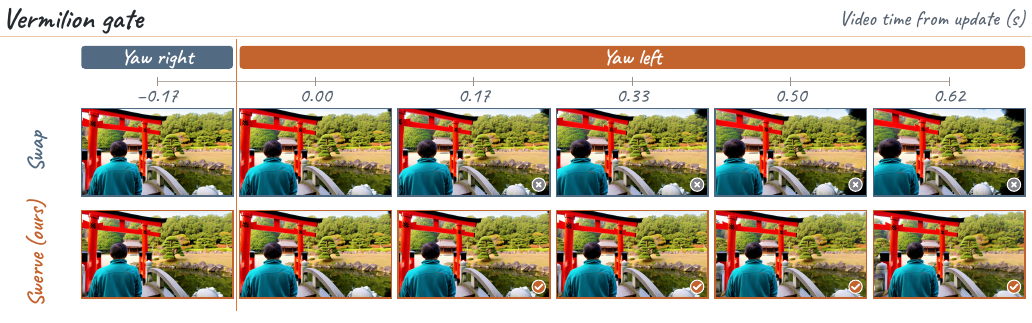}%
  \captionof{figure}{\textbf{Vermilion gate.} The action switches from yaw right to yaw left.}%
  \label{fig:supp-whole-vermilion-gate}%
\end{minipage}\par

\par\smallskip\noindent
\begin{minipage}{\linewidth}
  \centering
  \captionsetup{hypcap=false,skip=4pt}
  \includegraphics[width=.87\linewidth]{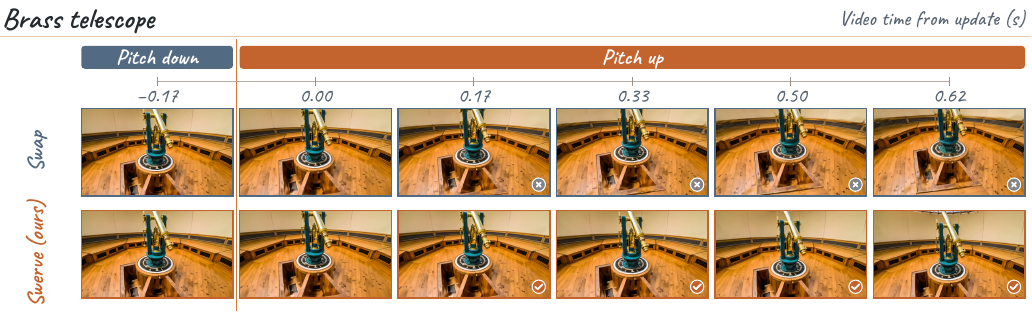}%
  \captionof{figure}{\textbf{Brass telescope.} The action switches from pitch down to pitch up.}%
  \label{fig:supp-whole-telescope}%
\end{minipage}\par

\par\smallskip\noindent
\begin{minipage}{\linewidth}
  \centering
  \captionsetup{hypcap=false,skip=4pt}
  \includegraphics[width=.87\linewidth]{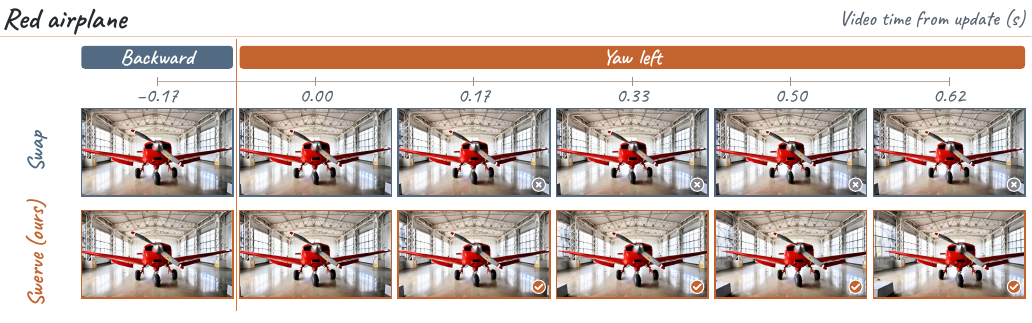}%
  \captionof{figure}{\textbf{Red airplane.} The action switches from backward to yaw left.}%
  \label{fig:supp-whole-airplane}%
\end{minipage}\par
\end{figure*}

\endgroup